\documentclass[11pt]{article}

\usepackage[T1]{fontenc}
\usepackage{lmodern}

\usepackage{fullpage}
\usepackage[utf8]{inputenc} 
\usepackage[T1]{fontenc}    
\usepackage{hyperref}       
\usepackage{url}            
\usepackage{booktabs}       
\usepackage{amsfonts}       
\usepackage{nicefrac}       
\usepackage{microtype}      
\usepackage{graphicx}
\usepackage{multirow}
\usepackage{booktabs}
\usepackage{color}
\usepackage{bm}
\usepackage{bbm}
\usepackage{amsmath}
\usepackage{amsthm}
\usepackage{amssymb}
\usepackage{xcolor} 
\usepackage{subfigure}
\usepackage{pdfpages}
\usepackage{natbib}

\usepackage{enumerate}
\usepackage{enumitem}

\usepackage{microtype}
\usepackage{graphicx}
\usepackage{subfigure}
\usepackage{booktabs} 
\usepackage{caption}
\usepackage{natbib}

\usepackage{enumerate}

\usepackage{amsmath}
\usepackage{amsthm}
\usepackage{amssymb}
\usepackage{thmtools,thm-restate}
\usepackage{amsfonts}
\usepackage{mathtools}
\usepackage{algorithm}
\usepackage{algorithmic}

\usepackage{enumitem}

\usepackage{tcolorbox}

\newtheorem{theorem}{Theorem}

\usepackage{mathtools}

\usepackage{chngcntr}
\usepackage{apptools}
\AtAppendix{\counterwithin{lemma}{section}}
\AtAppendix{\counterwithin{theorem}{section}}
\AtAppendix{\counterwithin{algorithm}{section}}
\AtAppendix{\counterwithin{figure}{section}}

\newcommand{\signSGD}{\textsc{signSGD}}

\newcommand{\medianGD}{\textsc{medianGD}}
\newcommand{\medianSGD}{\textsc{medianSGD}}

\def\Var{\mathrm{Var}}

\newtheorem{prop}{Proposition}

\definecolor{TC_color}{rgb}{0.858, 0.188, 0.778}
\definecolor{XC_color}{rgb}{0.1, 0.488, 0.478}
\definecolor{HS_color}{rgb}{0.36, 0.68, 0.68}

\usepackage[suppress]{color-edits}
\addauthor{sw}{blue}
\addauthor{mh}{red}
\addauthor{xc}{XC_color}
\addauthor{sun}{HS_color}

\allowdisplaybreaks[2]

\title{Distributed Training with Heterogeneous Data:\\ Bridging Median- and Mean-Based Algorithms}

\author{
  Xiangyi Chen\thanks{equal contribution} \thanks{University of Minnesota. Email: \texttt{chen5719@umn.edu}.}, ~ Tiancong Chen$^{*}$\thanks{University of Minnesota. Email: \texttt{chen6271@umn.edu}.}, ~ Haoran Sun\thanks{University of Minnesota. Email: \texttt{sun00111@umn.edu}.}, ~ Zhiwei Steven Wu\thanks{University of Minnesota. Email: \texttt{zsw@umn.edu}. Supported in part by a Google Faculty Research Award, a J.P. Morgan Faculty Award, and a Facebook Research Award.}, ~ Mingyi Hong\thanks{University of Minnesota. Email:  \texttt{mhong@umn.edu}.}
}
\date{}

\begin{document}

\maketitle

\begin{abstract}
Recently, there is a growing interest in the study of median-based algorithms for distributed non-convex optimization. Two prominent such algorithms include \signSGD{} with majority vote, an effective approach for communication reduction via 1-bit compression on the local gradients, and \medianSGD{}, an algorithm recently proposed to ensure robustness against Byzantine workers. The convergence analyses for these algorithms critically rely on the assumption that all the distributed data are drawn iid from the same distribution. However, in applications such as Federated Learning, the data across different nodes or machines can be inherently heterogeneous, which violates such an iid assumption.

This work analyzes \signSGD{} and \medianSGD{} in distributed settings with heterogeneous data.  
We  show that these algorithms are non-convergent whenever there is some disparity between the expected median and mean over the local gradients. To overcome this gap, we provide a novel gradient correction mechanism that perturbs the local gradients with noise, together with a series results that provable close the gap between mean and median of the gradients. The proposed methods largely preserve nice properties of these methods, such as the low per-iteration  communication complexity of \signSGD{}, and further enjoy global convergence to stationary solutions.
Our perturbation technique can be of independent interest when one wishes to estimate mean through a median estimator. 
\end{abstract}

\vspace{-0.1in}
\section{Introduction}
\vspace{-0.1in}
In the past few years, deep neural networks have achieved great successes in many tasks including computer vision and natural language processing. For many tasks in these fields, it may take weeks or even months to train a model due to the size of the model and training data set. One practical and promising way to reduce the training time of deep neural networks is using distributed training \citep{dean2012large}. A popular and practically successful paradigm for distributed training is the parameter server framework~\citep{paramserver}, where most of the computation is offloaded to workers in parallel and a parameter sever is used for coordinating the training process. Formally, the goal of such distributed optimization is to minimize the average of $M$ different functions from $M$ nodes, 
\vspace{-0.05in}
{\small \begin{align}
    \min_{x\in\mathbb{R}^d} f(x) \triangleq \frac{1}{M} \sum_{i=1}^M f_i(x),\label{formulation}
\end{align}}%
where each node $i$ can only access information of its local function $f_i(\cdot)$, defined by its local data.
\xcedit{Typically, such local objective takes the form of either an {\it expected} loss over local data distribution (population risk), or an {\it empirical average} over loss functions evaluated over finite number of data points (empirical risk). That is,
\vspace{-0.05in}
{\small\begin{align}\label{eq:local:function}
    f_i(x) = \int p_i(\zeta) l(x;\zeta) d\zeta, \quad\mbox{or}\quad f_i(x) =\frac{1}{K}\sum_{k=1}^{K}l(x;\zeta_{i,k})
\end{align}}%
where $l(x;\zeta)$ (resp. $l(x;\zeta_{i,k})$) is the cost evaluated at a given data point $\zeta$ (resp. $\zeta_{i,k}$).  
}

\mhedit{Similar to the parameter server paradigm,} motivated by the use of machine learning models on mobile devices, a distributed training framework called Federated Learning has become popular \citep{konevcny2016federated1,mcmahan2017communication,mcmahan2017federated}. In Federated Learning, the training data are \swedit{distributed across} personal devices and one wants to train a model without transmitting the users' data due to privacy concerns. 
\swedit{While many distributed algorithms proposed for parameter server are applicable to Federated Learning, Federated Learning posed many unique challenges, including the presence of \textit{heterogeneous} data across the nodes, and the need to accommodate asynchronous updates, as well as very limited message exchange among the nodes and the servers.} 
\xcedit{By {\it heterogeneous} data, we mean that either $p_i(\zeta)$, or the empirical distribution formed by $\{\zeta_{i,k}\}_{k=1}^{K}$ in \eqref{eq:local:function}, are significantly different across the local nodes. Clearly, when the data is heterogeneous, we will have $\nabla f_i(x)\ne \nabla f_j(x)$ 
and if local data are {\it homogeneous}, we will have $\nabla f_i(x) =  \nabla f_j(x)$ or  $\nabla f_i(x) \approx \nabla f_j(x)$ when $K$ is large. 
}

Recently, two prominent first-order algorithms under the parameter server framework are proposed. One is called \signSGD{} (with majority vote) (see Algorithm \ref{alg:signum}) \citep{bernstein2018signsgd}, \swedit{which updates} the parameters based on a majority vote of sign of gradients to reduce communication overheads. \xcedit{ The other one is called \medianGD{} (\mhedit{see its generalized stochastic version in Algorithm \ref{alg:mediangd}, which we refer to as \medianSGD{}} \citep{yin2018byzantine}), which aims to ensure robustness against Byzantine workers by using coordinate-wise median of gradients to evaluate mean of gradients.} \xcdelete{Note that in the table below, $g_{i,t}$ is the stochastic gradient at node $i$; the {sampling noise} comes from either drawing gradients from a given distribution, or from sub-sampling from finite number of local data set; see discussion from \eqref{eq:local:function} -- \eqref{eq:gradient:different}.} 

\begin{minipage}{0.45\textwidth}
\begin{algorithm}[H]
   \caption{\signSGD{} (with M nodes)}
   \label{alg:signum}
\begin{algorithmic}[1]
   \STATE {\bfseries Input:} learning rate $\delta$, current point $x_t$

   \STATE$g_{t,i}  \leftarrow \nabla f_i(x_t) + \textrm{sampling noise}$ 
   \STATE $x_{t+1} \leftarrow x_t - \delta \,\mathrm{sign} ( \sum_{i=1}^M \,\mathrm{sign}(g_{t,i}))$
\end{algorithmic}
\end{algorithm}
\end{minipage}
\begin{minipage}{0.45\textwidth}
\begin{algorithm}[H]
   \caption{\medianSGD{} (with M nodes)}
   \label{alg:mediangd}
\begin{algorithmic}[1]
   \STATE {\bfseries Input:} learning rate $\delta$, current point $x_t$
   \STATE  $g_{t,i}  \leftarrow \nabla f_i(x_t) + \textrm{sampling noise}$
   \STATE $x_{t+1} \leftarrow x_t - \delta \mathrm{median}(\{g_{t,i}\}_{i=1}^M)$
\end{algorithmic}
\end{algorithm}
\end{minipage}
\vspace{0.5cm}

 At first glance, \signSGD{} and \medianSGD{} seem to be two completely different algorithms with different update rules, designed for different desiderata (that is, communication-efficiency versus robustness). Surprisingly, we show that \signSGD{} can be viewed as updating variables along signed median direction ($\mathrm{sign}( \mathrm{median}(\{g_{t,i}\}_{i=1}^M))$), uncovering its hidden connection to \medianSGD{}.
 
 Currently, the analyses of both \signSGD{} and \medianSGD{} \xcedit{rely on the assumption of homogeneous data. }\xcdelete{rely on some form of iid assumptions on the local data across the nodes.} \signSGD{} is analyzed from the \swedit{in-sample} optimization perspective: it converges to stationary points if stochastic gradients $g_{t, i}$ sampled from each worker follow the same distribution \citep{bernstein2018signsgd,bernstein2019signsgd}. \xcedit{That is, at each given $x_t$, $\nabla f_i(x_t) = \nabla f_j(x_t)$, and the sampling noises follow the same distribution.}\swdelete{, which is equivalent to assuming all data are sampled iid.} \medianSGD{} is analyzed under the framework of population risk minimization:  it converges with an optimal statistical rate, but again under the assumption that the data across the workers are iid~\citep{yin2018byzantine}. \swdelete{Overall, both the convergence of \signSGD{} and \medianSGD{}/ \medianSGD{} require the data from different workers to be sampled from the same distribution.} \mhdelete{\mhedit{The iid assumption may be reasonable in parameter server based training}  as one can shuffle the data, then split and send the shuffled data to the workers, making the data distributions identical.} 
 
 However, in many modern distributed settings especially Federated Learning, data on different worker nodes can be inherently heterogeneous. 
 For example, users' data stored on different worker nodes might come from different geographic regions, which induce substantially different data distributions. In Federated Learning, the stochastic gradient $g_{t, i}$  from each device is effectively the full gradient $\nabla f_i(x_t)$ 
evaluated on the user's data 
\xcedit{(due to the small size of local data)}, which violates the assumption of identical gradient distributions. Therefore, under these heterogeneous data settings,  
 data aggregation and shuffling are often infeasible, and  there is very little understanding on the behavior of both median-based algorithms.
 
From the fixed point perspective, median-based algorithms like \signSGD{}, \medianSGD{}, and \medianSGD{} drive the median of gradients to 0\swedit{---that is, when median of gradients reaches 0, the algorithms will not perform updates}. When the median is close to the mean of gradients (the latter is the gradient of the target loss function), it follows that the true gradient is also approximately 0, \mhedit{and an approximate stationary solution is reached.} 
The reason of assuming homogeneous data in existing literature \citep{bernstein2018signsgd,bernstein2019signsgd,yin2018byzantine} is exactly to ensure that the median is close to mean. However, when the data from different workers are not drawn from the same distribution, the potential gap between the mean and median could prevent these algorithms from reducing the true gradient.\swdelete{there will be a gap between median and mean and one would expect when the median of gradients is driven to be 0 by the median-based algorithms, the true gradient can still be a constant.}

\begin{figure}[H]
\vspace{-0.5cm}
\centering
\subfigure[][Trajectory of \signSGD{}]{\label{fig:toy_signsgd} \includegraphics[height=2in]{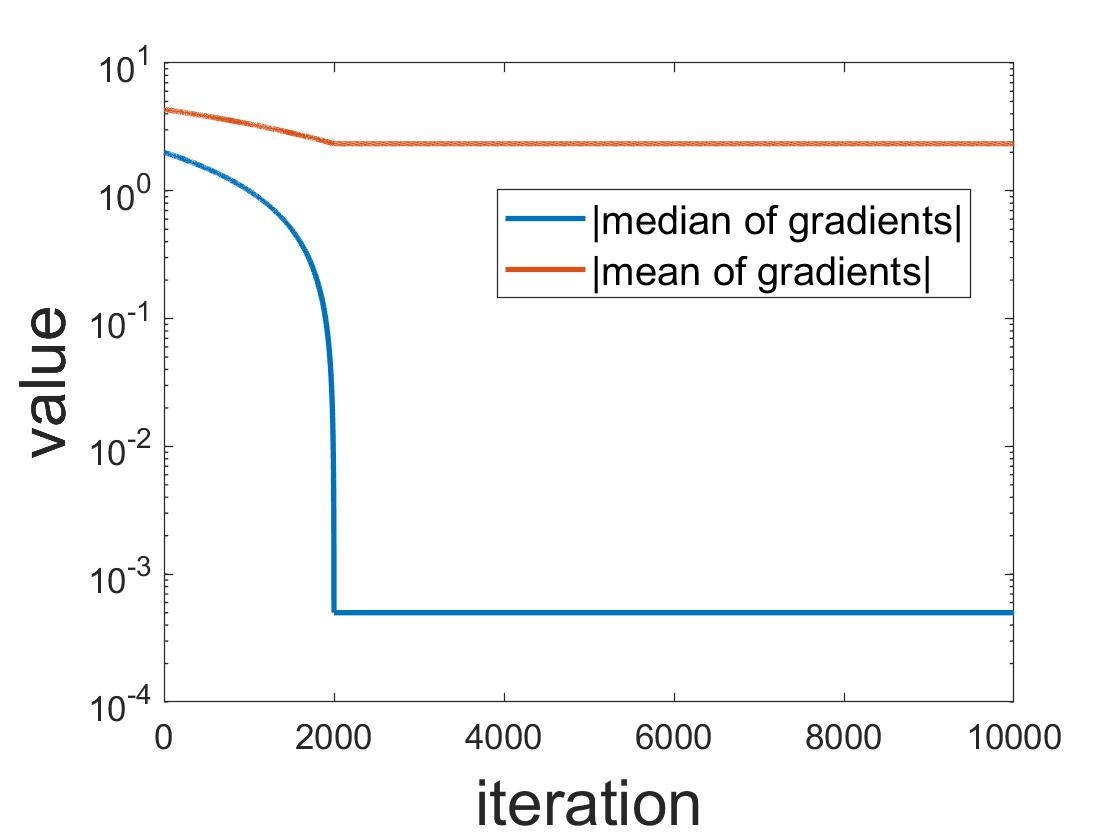}} \hspace{8pt}
\subfigure[][Trajectory of \medianSGD{}]{\label{fig:toy_mediangd} \includegraphics[height=2in]{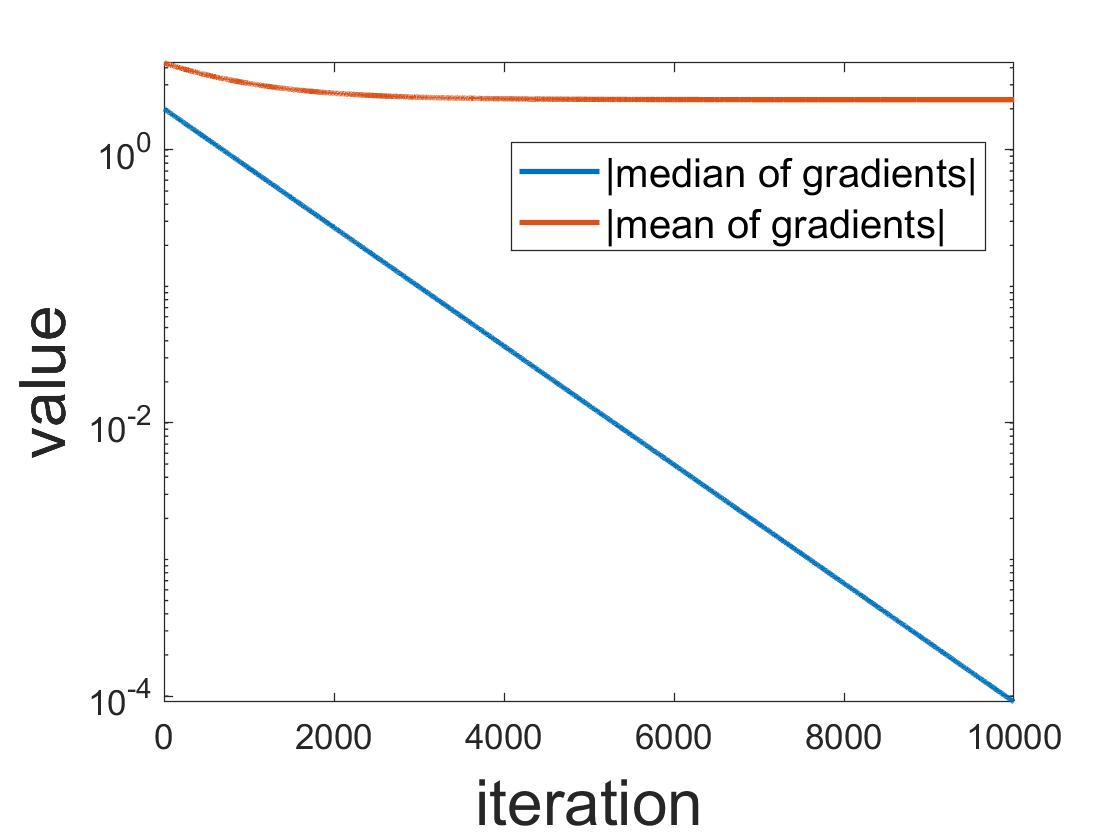}}
\caption[toyexample]{Absolute value of mean and median of gradient vs iteration. 
\subref{fig:toy_signsgd} shows the trajectory of \signSGD{}
\subref{fig:toy_mediangd} shows the trajectory of \medianSGD{} }\label{fig: diverge}
\vspace{-0.4cm}
\end{figure}

\swedit{To illustrate this phenomenon, consider a simple one-dimensional example: $\frac{1}{3}\sum_{i=1}^3 f_i(x) \triangleq (x-a_i)^2/2$, with 
$a_1 = 1, a_2 = 2, a_3 = 10$. If we run \signSGD{} and \medianSGD{}  with step size $\delta = 0.001$ and initial point $x_0 = 0.0005$, both algorithms  will produce iterates with large disparity between the mean and median gradients. See Fig. \ref{fig: diverge} for the trajectories of gradient norms. Both algorithms drive the median of gradients to 0 (\signSGD{} finally \swedit{converges to} the level of step size due to the use of sign in its update rule), while the true gradient remains a constant. In Sec. \ref{sec:exp}, we provide further empirical evaluation to demonstrate that such disparity can severely hinder the training performance.} \swdelete{The example implies that \signSGD{} and \medianSGD{} may not perform well when data on different nodes are not drawn iid. However, heterogeneous data appears in many existing applications where shuffle of data is prohibited. In addition, to use \signSGD{} and \medianSGD{} as general optimization algorithms instead of algorithms just for training neural nets, it is important to understand the behavior of these algorithms without assuming iid data.}


\noindent{\bf Our contribution.} Motivated by the need to understand median-based algorithms under heterogeneous data settings, we investigate two questions: 1)
in a distributed training environment, \mhdelete{with finite number of data samples per worker node,}
under what conditions do \signSGD{} and \medianSGD{} work well? and 2) can we provide mechanisms to \mhedit{close} the convergence gap in these algorithms?
Specifically, we analyze the convergence rate of \signSGD{} and \medianSGD{}, \mhedit{with finite number of data samples per worker node}, without assuming the data on different workers are from the same distribution. \mhdelete{In particular, we analyze the convergence of these algorithms under a finite-sum setting without any assumption on the underlying distribution of data.} Our contributions are summarized as following.

\begin{itemize}[leftmargin=*]
\vspace{-0.2cm}
   \item \textbf{\signSGD{} as a median-based algorithm}: We show that \signSGD{} actually updates variables along the direction of signed median of gradients, \mhedit{a fact that is crucial for its analysis, but has not been recognized  by existing literature so far.}
   
   \vspace{-0.15cm}
   \item  \textbf{Non-convergence due to the gap between median and mean.} We prove that \signSGD{} and \medianSGD{} usually cannot converge to stationary points in the limit. Instead, they are only guaranteed to find points \mhedit{whose gradient sizes are not vanishing, but in the order of the difference between expected median and mean of gradients at different workers}. 
   Further, we show that the non-convergence is not an artifact of analysis by providing examples where \signSGD{} and \medianSGD{} does not converge due to the gap between median and mean.
   
   \vspace{-0.15cm}
   \item \textbf{Bridging the gap between median and mean by adding controlled noise.} We prove the following key result:  given an arbitrary set of numbers, if one adds unimodal and symmetric noises \mhedit{with variance $\sigma^2$}, then the expected median of the resulting numbers will approach the expected mean of the original numbers, with a rate of $O(1/\sigma)$. 
   In addition, the distribution of the median will become increasingly symmetric as the variance of noise increases, with a rate of $O(1/\sigma^2)$. This result could be of independent interest. 
   
   \vspace{-0.15cm}
   \item \textbf{Convergence of noisy \signSGD{}  and noisy \medianSGD{}.} By using the fact that expected median converges to mean, and using a sharp analysis on the pdf of the noise on median, we prove that noisy \signSGD{} and noisy \medianSGD{} can both converge to stationary points. 
\end{itemize}
\vspace{-0.15cm}

\subsection{Related work} 

\paragraph{Distributed training and communication efficiency.} Distributed training of neural nets has become popular since the work of \citet{dean2012large},  in which distributed SGD was shown to achieve significant acceleration compared with SGD \citep{robbins1951stochastic}. As an example, \citet{goyal2017accurate} showed that distributed training of ResNet-50 \citep{he2016deep} can finish within an hour. There is a recent line of work providing methods for communication reduction in distributed training, including stochastic quantization~\citep{alistarh2017qsgd, wen2017terngrad} and 1-bit gradient compression such as \signSGD{}~\citep{bernstein2018signsgd,bernstein2019signsgd}

\paragraph{Byzantine robust optimization.} Byzantine robust optimization draws increasingly more attention in the past a few years. Its goal is to ensure performance of the optimization algorithms in the existence of Byzantine failures. \citet{alistarh2018byzantine} developed a variant of SGD based on detecting Byzantine nodes. \citet{yin2018byzantine} proposed \medianGD{} that is shown to converge with optimal statistical rate. \citet{blanchard2017machine} proposed a robust aggregation rule called Krum.  It is shown in \citet{bernstein2019signsgd} that \signSGD{} is also robust against certain failures.  Most existing works assume homogeneous data. In addition, \citet{bagdasaryan2018backdoor} showed that many existing Byzantine robust methods are vulnerable to adversarial attacks.

\paragraph{Federated Learning.} Federated Learning was initially introduced in \citet{konevcny2016federated1,mcmahan2017federated} for collaborative training of machine learning models without transmitting users' data. It is featured by high communication cost, requirements for failure tolerance and privacy protection, as the nodes are likely to be mobile devices such as cell phones. \citet{smith2017federated} proposed a learning framework that incorporates multi-task learning into Federated Learning. 
\citet{bonawitz2019towards} proposed a system design for large scale Federated Learning.

\paragraph{Non-convergence of \signSGD{}.} It is shown in \citet{reddi2019convergence} that Adam \citep{kingma2014adam} can diverge in some cases. Since \signSGD{} is a special case of Adam, it can suffer the same issue as Adam in general \citep{karimireddy2019error}. However, it is shown in \citet{bernstein2019signsgd} that when noise is unimodal and symmetric, \signSGD{} can guarantee convergence.

\vspace{-0.1in}
\section{Distributed \signSGD{} and \medianSGD{}}\label{sec: nonconverge}
\vspace{-0.1in}
In this section, we give convergence analyses of \signSGD{} and \medianSGD{} \swedit{for the problem defined in \eqref{formulation}}, without any assumption on data distribution. \mhedit{All proofs of the results can be found in Appendix \ref{app: proof_prop_1} -- \ref{app: proof_thm2}.} \swdelete{Throughout the paper, we consider the classical distributed formulation, which minimizes the average of $M$ different functions from $M$ nodes. Each node can only access information of its local function, which is smooth and possibly nonconvex. The objective can be written by
\begin{align}
    \min_{x\in\mathbb{R}^d} f(x) \triangleq \frac{1}{M} \sum_{i=1}^M f_i(x).
\end{align}
 In empirical risk minimization, $f_i$ can be viewed as the average loss evaluated on data at $i$th worker.
}
We first analyze the convergence of the algorithms under the framework of stochastic optimization. In such a setting, at iteration $t$, worker $i$ can access a stochastic gradient estimator $g_i(x_t)$ (also denoted as $g_{t,i}$ for simplicity).  Denote the collection of the stochastic gradients to be $\{g_t\}$, we make following assumptions for the analysis.

\textbf{A1:} The stochastic gradients obtained from different nodes are unbiased 
{\small \begin{align}
    \mathbb E[g_{i}(x)] = \nabla f_i(x)
\end{align}}%
\textbf{A2:} The median of gradient estimators have bounded variance on each coordinate, i.e. $\forall j \in [d]$,
{\small \begin{align}
    \mathbb E[\| \mathrm{median}(\{g_t\})_j - \mathbb E[ \mathrm{median}(\{g_t\})_j|x_t]\|^2] \leq \sigma_m^2 
\end{align}}%
\textbf{A3:} $f$ has Lipschitz gradient, i.e.
{\small \begin{align}
    \|\nabla f(x) - \nabla f(y)\| \leq L \|x-y\|
\end{align}}%
\textbf{A4:}  $M$ is an odd number.

The assumptions A1 and A3 are standard for stochastic optimization. A4 is just to ensure that \signSGD{} can be rigorously viewed as a median-based algorithm (see Proposition \ref{prop: sign_med}).  In stochastic optimization, it is usually also needed to assume the gradient estimators have bounded variance. A2 is a variant of bounded variance for median-based algorithms.   It is almost equivalent to assuming bounded variance on the gradient estimator on each node, since the variance on the median will be in the same order as the maximum variance on each individual node. 

\textbf{Notations}: Given a set of vectors $a_i,\, i = 1,...,n$, we denote $\{a_{i}\}_{i=1}^n$ to be the the set and $\mathrm{median}(\{a_{i}\}_{i=1}^n)$ to be the coordinate-wise median of of the vectors. We also use $\mathrm{median}(\{a\})$ to denote $\mathrm{median}(\{a_{i}\}_{i=1}^n)$ for simplicity. When $v$ is a vector and $b$ is a constant, $v \neq b$ means none of the coordinate of $v$ equals $b$. Finally, $(v)_j$ denotes $j$th coordinate of $v$, $\mathrm{sign}(v)$ denotes the signed vector of $v$. \mhedit{We use [N] to denote the set $\{1,2,\cdots, N\}$.}
\vspace{-0.05in}
\subsection{Convergence of \signSGD{} and \medianSGD{}}
\vspace{-0.05in}
From the pseudo code of Algorithm 1, it is not straightforward to see how \signSGD{} is related to median of gradients, since there is no explicit operation for calculating median in the update rule of \signSGD{}. It turns out that \signSGD{} actually goes along the signed median direction, which is given by Proposition \ref{prop: sign_med}. 
\begin{prop} \label{prop: sign_med}
When $M$ is odd and $\mathrm{median}(\{g_{t}\}) \neq 0$, we have 
{\small \begin{align}
  \mathrm{sign} ( \sum_{i=1}^M \,\mathrm{sign}(g_{t,i}))   =  \mathrm{sign}( \mathrm{median}(\{g_{t}\}))
\end{align}}%
\end{prop}
Thus, \signSGD{} updates the variables base on sign of coordinate-wise median of gradients, while \medianSGD{} updates the variables toward the median direction of gradients. Though these two algorithms are connected to each other, none of their behavior is well-understood. We provide the convergence guarantee for these algorithms in Theorem \ref{thm: signsgd} and Theorem \ref{thm: median_gd}, respectively.

\begin{theorem}\label{thm: signsgd}
Suppose A1-A4 are satisfied, 
and define $D_f \triangleq f(1) - \min_{x}f(x)$. For \signSGD{} with $\delta = \frac{\sqrt{D_f}}{\sqrt{LdT}}$, the following holds true
{\small\begin{align} \label{eq: rate_gradient}
    \frac{1}{T} \sum_{t=1}^T \mathbb E[ \| \nabla f(x_t)\|_1]    
     \leq  \frac{3}{2} \frac{\sqrt{dLD_f}}{\sqrt{T}} + 2 \frac{1}{T }\sum_{t=1}^T \mathbb E[\|\mathbb E[\mathrm{median}(\{g_t\})]|x_t] - \nabla f(x_t)\|_1]  
      + 2 d \sigma_m  
\end{align}}%
\end{theorem}
\mhedit{One key observation from Theorem \ref{thm: signsgd} is that,} as $T$ goes to infinity, the RHS of \eqref{eq: rate_gradient} is dominated by the difference between median and mean and the standard deviation on the median.

We remark that under the assumption that the gradient estimators from different nodes \xcedit{are drawn from the same distribution in \signSGD{},} \xcdelete{have the same mean and unimodal and symmetric noises in \signSGD{},} the analysis recovers the bound in \citet{bernstein2018signsgd}. \mhedit{This is because when the mean of the gradient esimators are the same, we have:} 
{\small$${\mathbb E[\mathrm{median}(\{g_t\})|x_t] = \mathbb E[\nabla f(x_t)]}.$$ }%
\xcedit{Further, if the noise on each coordinate of local gradients has variance bounded by $\sigma_l^2$, then $\sigma_m   = O(\sigma_l/\sqrt{M})$ (see Theorem 1.4.1 in \citet{miller2017probability}). 
Thus \eqref{eq: rate_gradient} becomes
{\small\begin{align} \label{eq: rate}
    \frac{1}{T} \sum_{t=1}^T \mathbb E[ \| \nabla f(x_t)] \|_1 
     \leq  \frac{3}{2} \frac{\sqrt{dLD_f}}{\sqrt{T}} 
      + d O(\frac{\sigma_l}{ \sqrt{M}})
\end{align}}%
which recovers the convergence rate in \citet{bernstein2018signsgd}.} Under minibatch setting, we can further set the minibatch size to be $O(\sqrt{T})$ to make $\sigma_l$ \mhedit{proportional to} $O(1/\sqrt{T})$. 

\begin{theorem}\label{thm: median_gd}
Suppose A1-A4 are satisfied, define $D_f \triangleq f(1) - \min_{x}f(x)$.  Assume $|\mathbb E[\mathrm{median}(\{g_t\})_j|x_t] - \nabla f(x_t)_j| \leq C,\, \forall j \in [d]$, set $\delta = \min(\frac{1}{\sqrt{Td}},\frac{1}{3L})$, 
\medianSGD{} yields
{\small\begin{align}\label{eq: rate_median_gd}
    \frac{1}{T}\sum_{t=1}^T  \mathbb E[ \|\nabla f(x_t)\|^2] 
     \leq & \frac{2\sqrt{d}}{\sqrt{T}} D_f  + {3L} \frac{\sqrt{d}}{\sqrt{T}} (\frac{\sigma_m^2}{d} + C^2 ) +  \frac{2}{T}\sum_{t=1}^T \mathbb E[\|\nabla f(x_t) - \mathbb E[\mathrm{median}(\{g_t\})|x_t]\|^2]   \nonumber \\
     &+  \frac{2}{T}\sum_{t=1}^T \mathbb E[\| \mathbb E[\mathrm{median}(\{g_t\})|x_t]\|\|\nabla f(x_t) - \mathbb E[\mathrm{median}(\{g_t\})|x_t]\|] 
\end{align}}%
\end{theorem}
As $T$ increases, the RHS of \eqref{eq: rate_median_gd} will be dominated by the terms involving the difference between expected median of gradients and the true gradients. In the case where the gradient from each node follows the same symmetric and unimodal distribution, the difference vanishes and the algorithm converges to a stationary point with a rate of $\frac{\sqrt{d}}{\sqrt{T}}$.

\mhedit{However, when the gap between expected median of gradients and the true gradients is not zero, our results suggest that both \signSGD{} and \medianSGD{} can only converge to solutions where the size of the gradient is upper bounded by some constant related to the median-mean gap.} 
\vspace{-0.05in}
\subsection{Tightness of the convergence analysis}
\vspace{-0.05in}
\mhedit{Our convergence analysis suggests that it is difficult to make the {\it upper bounds} on the average size of the gradient of \signSGD{} and \medianSGD{} go to zero. However, it is by no means clear whether this gap is an artifact of our analysis, or it indeed exists in practice.}


We answer this question by providing examples to demonstrate that such convergence gap indeed exists, thus showing that the gap in the convergence analysis is inevitable unless additional assumptions are enforced. The proof for the results below are given in Appendix \ref{app: proof_thm3}.

\begin{theorem}\label{thm: diverge_signsgd}
There exists a problem instance where \signSGD{} converges to a point $\hat{x}^*$ with 
{\small\begin{align}
    \|\nabla f(\hat{x}^*)\|_1 \geq   \frac{1}{T }\sum_{t=1}^T \mathbb E[\|\mathbb E[\mathrm{median}(\{g_t\})]] - \nabla f(x_t)\|_1]  \geq 1.
\end{align}}
and \medianSGD{} converges to
{\small\begin{align}
 \|\nabla f(\hat{x}^*)\|^2  
 \geq & \frac{1}{T }\sum_{t=1}^T \mathbb E[\|\mathbb E[\mathrm{median}(\{g_t\})]] - \nabla f(x_t)\|^2]
     \geq 1.
\end{align}}
\end{theorem}


At this point, it appears to be hopeless to make \signSGD{} and \medianSGD{} converge to stationary solutions, unless assuming the expected median of gradients is close to the mean, as has been done in many existing works \citep{bernstein2018signsgd,bernstein2019signsgd,yin2018byzantine}. Unfortunately, this assumption may not be valid for important applications such as Federated Learning, where the data  located at different workers are not  i.i.d. and they cannot be easily shuffled. 
\vspace{-0.1in}
\section{Convergence of median towards the mean}\label{sec: bridge_gap}
\vspace{-0.1in}
In the previous section, we saw that there could be a convergence gap depending on the difference between expected median and mean for either \signSGD{} or \medianSGD{}. \mhdelete{A natural solution to the problem is to make expected median to be the same as mean. Then the question is how to achieve it without shuffling the data.}

In the following, we present a general result showing that the expected median and the mean can be closer to each other, if some random noise is properly added. This is the key leading to  our proposed perturbation mechanism, which ensures \signSGD{} and \medianSGD{} can converge properly.


\begin{theorem}\label{thm: gap_median_mean}
    Assume we have a set of numbers $u_1,..,u_{2n+1}$. Given a symmetric and unimodal noise distribution with mean 0, variance 1. Denote the pdf of the distribution to be $h_0(z)$ and cdf to be $H_0(z)$. Suppose $h_0'(z)$ is uniformly bounded and absolutely integrable.  Draw $2n+1$ samples $\xi_1,...,\xi_{2n+1}$ from the distribution $h_0(z)$. Define random variable $\hat{u}_i = u_i + b\xi_i$, 
    
(a) We have
\vspace{-0.1in}
{\small\begin{align}\label{eq: scale_gap}
    \mathbb E[\mathrm{median}(\{\hat{u}_i\}_{i=1}^{2n+1})] = \frac{1}{2n+1}\sum_{i=1}^{2n+1}u_i +  O\left(\frac{\max_{i,j}{|u_i-u_j|^2}}{b}\right),
\end{align}}
\vspace{-0.1in}
{\small\begin{align}\label{eq: variance_scale}
    \Var(\mathrm{median}(\{\hat{u}_i\}_{i=1}^{2n+1}))=  O(b^2).
\end{align}}%
(b) Further assume $h_0''(z)$ is uniformly bounded and absolutely integrable.  Denote $r_b(z)$ to be the pdf of the distribution of $\mathrm{median}(\{\hat{u}_i\}_{i=1}^{2n+1})$ and $\bar{u} \triangleq \frac{1}{2n+1}\sum_{i=1}^{2n+1}u_i$, we have\mhedit{
\begin{align}
    r_b(\bar{u}+z) & = \underbrace{\frac{1}{b}g(\frac{z}{b})}_{\mbox{\rm symmetric part}} + \underbrace{\frac{1}{b}v(\frac{z}{b})}_{\mbox{\rm asymmetric part}}\label{eq:split_pdf}\\
 \mbox{where} \; \;  g(z) & \triangleq  \sum_{i=1}^{2n+1}  h_0(z) \sum_{S \in \mathcal{S}_i} \prod_{j \in S}  H_0(z)  \prod_{k \in [n]\setminus \{i,S\}} H_0(-z)  \label{eq: pdf_symmetric}
\end{align}}%
being the pdf of sample median of $2n+1$ samples drawn from distribution $h_0(z)$ which is symmetric over 0, \mhedit{and the asymmetric part satisfies}
{\small\begin{align}\label{eq:asymtc}
    \int_{-\infty}^{\infty} 
\frac{1}{b}|v(\frac{z}{b})|dz =  O\left(\frac{\max_i |\bar{u}- u_i|^2}{b^2} \right) 
\end{align}}
\vspace{-0.1in}
\end{theorem}
Eq. \eqref{eq: scale_gap} is one of the key results of Theorem \ref{thm: gap_median_mean}, i.e., the difference between the expected median and mean shrinks with a rate of $O(1/b)$ as $b$ grows. 
Another key result of the distribution is \eqref{eq:asymtc}, which says the pdf of the expected median becomes increasingly symmetric and the asymmetric part diminishes with a rate of $O(1/b^2)$. \mhdelete{as the variance of the added noise grows.} On the other hand, Eq \eqref{eq: variance_scale} is somewhat expected since the variance on the median should be at the same order as individual variance, which is $O(b^2)$.

It is worth mentioning that Gaussian distribution satisfies all the assumptions in Theorem \ref{thm: gap_median_mean}. In addition, although the theorem is based on assumptions on the second order differentiability of the pdf, we observe \mhedit{empirically} that, many commonly used symmetric distribution with non-differentiable points such as Laplace distribution and uniform distribution can also \mhdelete{have the key properties in the theorem. That is, they can} make the pdf increasingly symmetric and make the expected median closer to the mean, as $b$ increases.



\vspace{-0.1in}
\section{Convergence of Noisy \signSGD{} and Noisy \medianSGD{}}
\vspace{-0.1in}
From Sec. \ref{sec: bridge_gap}, we see that the gap between expected median and  mean will be reduced if sufficient noise is added. Meanwhile, from the analysis in Sec. \ref{sec: nonconverge}, \medianSGD{} and \signSGD{} will finally converge to some solution whose gradient size is proportional to the above median-mean gap \mhdelete{difference between expected median and mean of the gradients.}. 
\mhedit{Can we combine these results, so that we inject noise to the (stochastic) gradients 
to improve the convergence properties of median-based methods? The answer  is not immediately clear, 
since for this approach to work, the injected noise has to be {\it large enough} so that the mean-media gap is reduced quickly, while it has to be {\it small enough} to ensure that the algorithm still converges to a good solution.}   

\mhedit{In this section, we propose and analyze the following variants of \signSGD{} and \medianSGD{}, where symmetric and unimodal noises are injected on the local gradients. By performing a careful analysis, we identify the correct amount of noise to be injected, so that the algorithm converges with a suitable rate to any prescribed size of gradient. 
} \mhcomment{is the last sentence correct?}

\begin{minipage}{0.45\textwidth}
\begin{algorithm}[H]
   \caption{Noisy \signSGD{}}
   \label{alg:signsgdn}
\begin{algorithmic}
   \STATE {\bfseries Input:} learning rate $\delta$, current point $x_t$

   \STATE $g_{t,i} = \nabla f_i(x_t) + b \xi_{t,i} $   
   \STATE $x_{t+1} \leftarrow x_t - \delta \,\mathrm{sign} ( \sum_{i=1}^M \,\mathrm{sign}(g_{t,i}))$
\end{algorithmic}
\end{algorithm}
\end{minipage}
\begin{minipage}{0.45\textwidth}
\begin{algorithm}[H]
   \caption{Noisy \medianSGD{}}
   \label{alg:mediangdn}
\begin{algorithmic}
   \STATE {\bfseries Input:} learning rate $\delta$, current point $x_t$

   \STATE $g_{t,i} = \nabla f_i(x_t) + b \xi_{t,i} $ 
   \STATE $x_{t+1} \leftarrow x_t - \delta \mathrm{median}(\{g_{t,i}\}_{i=1}^M)$
\end{algorithmic}
\end{algorithm}
\end{minipage}
\vspace{0.5cm}



\textbf{Remark on the source of noise:} The above algorithms still follow the update rule of \signSGD{} and \medianSGD{}, just that the noises on the gradients follows some distributions with good properties described in Section \ref{sec: bridge_gap}. The easiest way to realize such noise is to let workers artificially inject the noise on their gradients. \mhedit{In machine learning applications, a more natural way to inject the noise is through data sub-sampling} when evaluating the gradient (\signSGD{} is already doing this!). It is shown in \citet{bernstein2018signsgd} that the noise on gradient obtained through sub-sampling is approximate unimodal and symmetric which is the main assumption in Theorem \ref{thm: gap_median_mean}. Intuitively, approximate unimodal and symmetric noise may also help reduce the gap between expected median and mean of gradients. Though this is not a rigorous argument, we show later in Section \ref{sec: experiments} that sub-sampling indeed help in practice under the situation of \swedit{heterogeneous} data. \xcedit{Our analysis assumes each worker evaluates the local \emph{full} gradients $\nabla f_i(x_t)$ for simplicity,  and we provide a discussion on how to prove convergence after combining perturbation with sub-sampling in Section \ref{sec: discuss}. Note that in our main application of Federated Learning, one usually computes $\nabla f_i(x_t)$ due to the small size of local data.} \swcomment{please check}\mhcomment{or we can say that, due to the size of the local data is small, it is relatively easy to compute full local gradient then inject the desired noise.}\mhcomment{where is the constant $Q$ used, and do we need the size of the per-coordinate gradient to be bounded?}

Theorem \ref{thm: noisy_signsgd} and Theorem \ref{thm: noisy_median_gd} provides convergence rates for Noisy \signSGD{} and Noisy \medianSGD{}, when the standard deviation of the noise is chosen to be $O(T^{1/4})$. 

\begin{theorem}\label{thm: noisy_signsgd}
Suppose A1, A3, A4 are satisfied and $|\nabla f_i(x_t)_j| \leq Q, \forall t,j$. When each $\xi_{t,i}$ is sampled iid from a symmetric and unimodal distribution with mean 0, variance 1 and pdf $h(z)$. If $h'(z)$ and $h''(z)$ are uniformly bounded and absolutely integrable. Set $b = T^{1/4}d^{1/4}$ and $\delta = 1/\sqrt{Td}$. Define $\sigma$ to be standard deviation of median of $2n+1$ samples drawn from $h(z)$, $
\mathcal W_t$ to be the set of coordinates $j$ at iteration $t$ with $\frac{|\nabla f(x_t)_j|   }{b\sigma} \geq \frac{2}{\sqrt{3}}$,  we have the following estimate
\vspace{-0.1in}
{\small\begin{align} \label{eq: rate_signsgd_n}
    &  \frac{1}{T}\sum_{t=1}^T\bigg( \sum_{j \in \mathcal W_t}T^{1/4}d^{1/4}|\nabla f(x_t)_j| + \sum_{j \in [d] \setminus \mathcal W_t} \nabla f(x_t)_j^2\bigg)
    \leq O\left(\frac{d^{3/4}}{T^{1/4}}\right)
\end{align}}
\vspace{-0.3in}
\end{theorem}
Theorem \ref{thm: noisy_signsgd} indicates the large coordinates of the gradient will shrink quite fast with a rate of $\frac{\sqrt{d}}{\sqrt{T}}$, until it is small compared with the variance. Note that the variance of the injected noise will be quite large if $T$ or $d$ is large enough. In this case, $\sigma$ is large, $\mathcal W_t$ will be mostly empty and the convergence measure of \signSGD{} become the classical squared $L_2$ norm of gradients. It converges with a rate of $O(d^{3/4}/T^{1/4})$ which is $O(T^{1/4}d^{1/4})$ slower compared with SGD (which has a rate of $O(\sqrt{d}/\sqrt{T})$ when assuming coordinate-wise bounded variance).

\begin{theorem}\label{thm: noisy_median_gd}
Suppose A1, A3, A4 are satisfied and  $|\nabla f_i(x_t)_j| \leq Q, \forall t,j$. When each coordinate of $\xi_{t,i}$ is sampled iid from a symmetric and unimodal distribution with mean 0, variance 1 and pdf $h(z)$. If $h'(z)$ is uniformly bounded and absolutely integrable.  Set $b = T^{1/4}d^{1/4}$ and $\delta = 1/\sqrt{Td}$,  pick $R$ uniformly randomly from 1 to $T$, for \medianSGD{}, we have
\vspace{-0.05in}
{\small\begin{align} \label{eq: rate_mediangd_n}
    &  E\left[ \|\nabla f(x_R)\|^2\right]
    \leq O\left(\frac{d^{3/4}}{T^{1/4}} \right)
\end{align}}%
\vspace{-0.3in}
\end{theorem}
Theorem \ref{thm: noisy_median_gd} shows that under suitable parameter setting, \medianSGD{} converges with the same rate as \signSGD{}. This is not surprising since both are median-based algorithms, and the diminishing speed of the gap between median and mean is controlled by how much noise is added.  

\textbf{Remark on the amount of noise:} In Theorem \ref{thm: noisy_signsgd} and Theorem \ref{thm: noisy_median_gd}, the variance of the noise must depend on $T$ and goes to $\infty$ with $T$ to guarantee convergence to stationary points in the limit. This is because a constant variance can only reduce the gap between expected median and mean by a constant factor, guaranteeing a small but non-zero gradient, which is not strong enough to be a stationary point. However, in practice, a constant variance is usually enough for problems such as neural network training.  This is due to at least following three reasons. i). The gap usually shrinks faster than the rates in Theorem \ref{thm: gap_median_mean}, since Theorem \ref{thm: gap_median_mean} is a worst case analysis. ii). For over-parameterized neural nets, the gradients on each data point are usually small at local minima when every data point is well-classified, implying a small gap between expected median and mean of gradients around local minima. iii). Early stopping is usually used to prevent overfiting, therefore optimization algorithms will stop when the gradient is small instead of being exactly zero. 
\vspace{-0.1in}
\section{Experiments}\label{sec: experiments}\label{sec:exp}
\vspace{-0.1in}
In this section, we show how noise helps the practical behavior of the algorithm. Since \signSGD{} is better studied empirically and \medianSGD{} is more of theoretical interest so far, we use \signSGD{} to demonstrate the benefit of injecting noise. {Our distributed experiments are implemented via Message Passing Interface (MPI) and related to training two-layer neural networks using the MNIST data set. The data distribution on each node is heterogeneous, i.e., each node only contains exclusive data for one or two out of ten categories. } 

\begin{figure}[H]
\centering
\vspace{-0.3cm}
\includegraphics[height=1.6in]{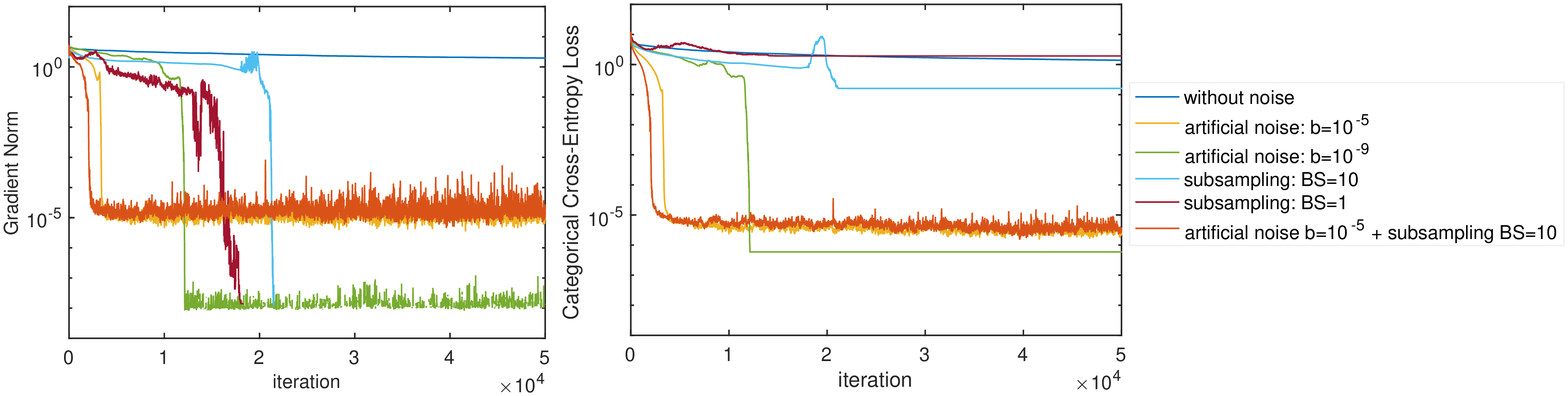}
\vspace{-0.3cm}
\caption[MNIST]{\small MNIST comparison of \signSGD{} with/without noise. \sundelete{Two kinds of noise are considered, 1) artificial random noise with zero mean and constant standard deviation b and 2) subsampling with specific batch size BS.} We train a two-layer fully connected neural network \sundelete{(with 128 and 10 neurons for each layer)} on a small subset of MNIST distributed over 5 machines, each machine has 100 data points. 
\sundelete{Left figure shows the dynamic of norm of gradient, while right figure shows the dynamic of categorical cross entropy loss.}
\xcdelete{We can observe that injecting noise does help in optimization.} \sundelete{Compared to the noiseless case, the gradients in noisy cases can converge to better solutions and oscillate at the corresponding noise standard deviation (i.e., $10^{-5}$ and $10^{-9}$), while the gradient error of the noiseless case is dominated by the gap of the difference between median and mean. We can also observe that, although the sub-sampling can reduce the gradient error, the performance on categorical cross entropy loss is much worse than injecting noise. Adding extra artificial noise to the subsampling also helps improve the performance. Detailed choice of algorithm parameters are given in  Sec. \ref{Implementation_Detail}.}} 
\label{fig:MNIST1}
 \vspace{-0.5cm}
\end{figure}

We first study the asymptotic performance of different algorithms, where we use a subset of MNIST and train neural networks until convergence.
\sundelete{Fig. \ref{fig:MNIST1} {studies the asymptotic performance when algorithm runs till convergence and} shows the evolution of gradient norm {and categorical cross entropy loss}.}  {We compare Noisy \signSGD{} (Algorithm \ref{alg:signsgdn}) with different $b$ (i.e. adding random Gaussian noise  with different standard deviation)}, 
 \signSGD{} with sub-sampling on data, and \signSGD{} without any noise.  We see that  \signSGD{} without noise stuck at some point where gradient is a constant (the median should be oscillating around zero). At the same time, both Noisy \signSGD{} or \signSGD{} with sub-sampling drive the gradient to be small. It should be noticed that {the \signSGD{} without noise converges to solutions where the sizes of the gradients are quite large}, compared with the amount of noise added by Noisy \signSGD{} or \signSGD{} with sub-sampling. Thus, the exploration effect of the noise may also contribute to making the final gradient small\xcedit{, since the noise added is not strong enough to bridge the gap}. \sundelete{Also, decreasing noise gradually seems to be a good strategy for training neural nets. The reason might be that at initial stage of training, the gradients are large and the gap between median and mean is large, a large noise help the algorithm go along the direction of gradient. As a local minimum is approached, the gradients from different workers are all small and thus the gap between median and mean is also small, thus a small noise is sufficient. }

\begin{figure}[H]
\centering
\vspace{-0.5cm}
\subfigure[][Norm of Gradient]{\label{fig:MNIST1-a} \includegraphics[height=1.5in]{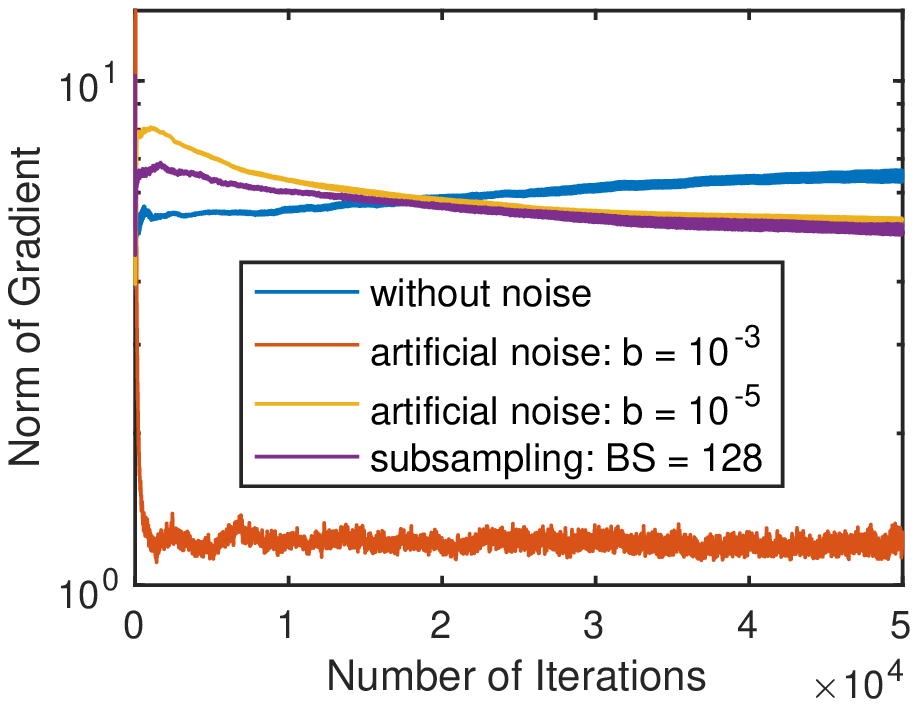}} 
\subfigure[][Training Accuracy]{\label{fig:MNIST1-c} \includegraphics[height=1.5in]{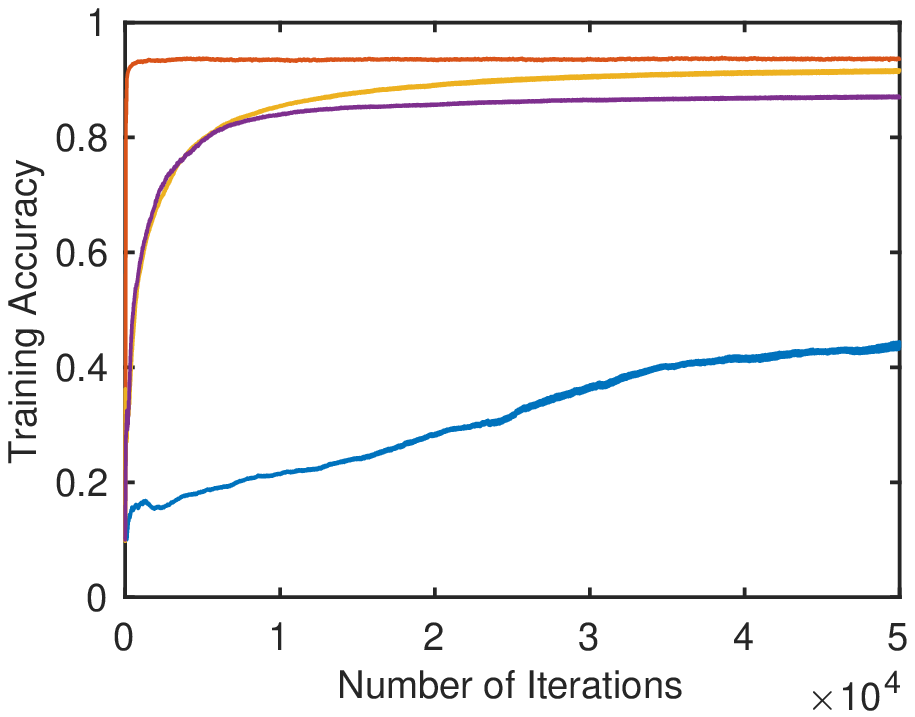}} 
\subfigure[][Testing Accuracy]{\label{fig:MNIST1-d} \includegraphics[height=1.5in]{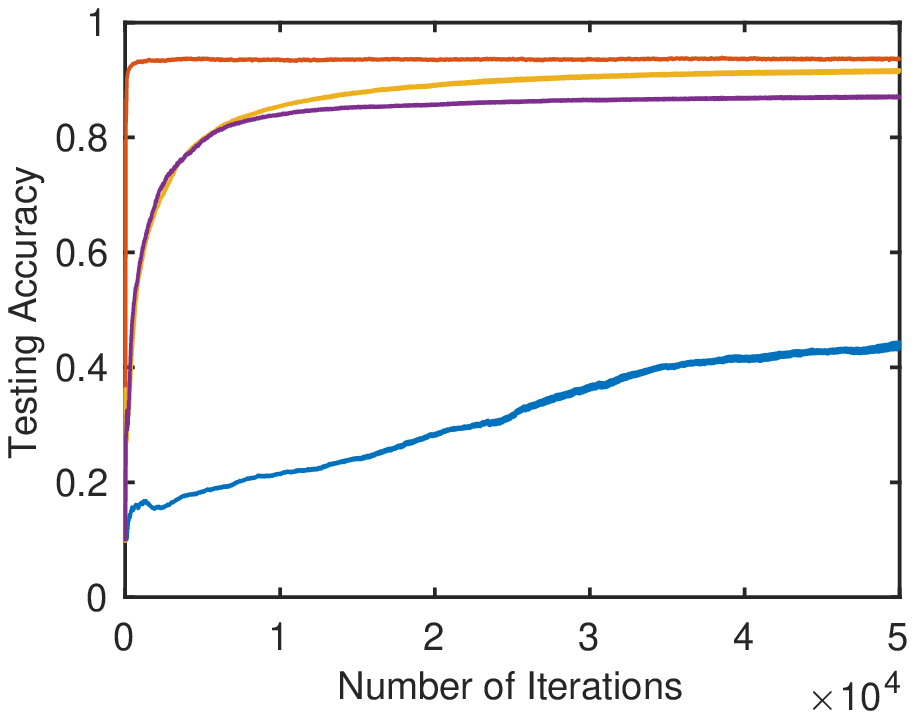}}
\vspace{-0.3cm}
\caption[MNIST]{\small MNIST comparison of \signSGD{} with and without noise. We train a two-layer fully connected neural network \sundelete{(with 128 and 10 neurons for each layer)} on MNIST dataset distributed over 10 machines. \sundelete{, each machine contains all training data from one of the ten classes (total 60,000 data points over 10 machines). The testing performance is evaluated on the test set of 10,000 data points. Four different metrics are shown here, in which \subref{fig:MNIST1-a}  Norm of gradient; \subref{fig:MNIST1-b} Training loss \subref{fig:MNIST1-c} Training accuracy; and, \subref{fig:MNIST1-d} Testing accuracy.} \xcdelete{We can observe that injecting noise does help improve training and testing performance. It is also more favorable to add artificial noise than subsampling.}}
\label{fig:MNIST2}
\end{figure}

\xcedit{In the second experiment, we examine the performance of Noisy \signSGD{} in a more realistic setting, where the full MNSIT dataset is used, and the training/testing accuracies are also compared.}
As mentioned in the previous section, early stopping is usually used for preventing over fitting and the optimization stops when the gradient is a constant. From this perspective, the perturbation may not be necessary since the gap between median and mean may be tolerable. However, from the results in Figure \ref{fig:MNIST2}, we can see that both training accuracy and test accuracy of \signSGD{} without noise are very poor. The perturbation indeed help improve the classification accuracy in practice.

\vspace{-0.3in}
\section{Conclusion and discussion}\label{sec: discuss}
\vspace{-0.1in}
In this paper, we uncover the connection between \signSGD{} and \medianSGD{} by showing \signSGD{} is a median-based algorithm. We also show that when the data at different nodes come from different distributions, \mhedit{the class of median-based algorithms suffers} from non-convergence caused by using median to evaluate mean. To fix the non-convergence issue, We provide a perturbation mechanism to shrink the gap between expected median and mean. After incorporating the perturbation mechanism into \signSGD{} and \medianSGD{}, we show that both algorithms can guarantee convergence to stationary points with a rate of $O({d^{3/4}}/{T^{1/4}})$. 
\mhedit{To the best of our knowledge, this is the first time that median-based methods, including \signSGD{} and \medianSGD{}, are able to converge with provable rate for distributed problems with heterogeneous data.} The perturbation mechanism can be approximately realized by sub-sampling of data during gradient evaluation, which partly support the use of sub-sampling in practice. We also conducted experiments on training neural nets to show the necessity of the perturbation mechanism and sub-sampling.

\mhedit{After all the analysis in the paper, several open questions remain, as we discuss them below. We plan to address these questions in our future works.}

\noindent{\bf Combining the perturbation  with sub-sampling?} 
\mhedit{We believe that it should be possible to combine the proposed perturbation technique with the sub-sampling technique used in training, as the expectation of average of sub-sampled gradients from different workers is the true gradient.} If one dive into the analysis, one may notice that the analysis of \medianSGD{} depends on characterizing the scaling of variance on median, while that for \signSGD{} depends on bounding the asymmetricity of pdf of median. For \medianSGD{}, the variance will not change much with sub-sampling, since sub-sampling creates at most a constant variance.  For \signSGD{}, bounding asymmetricity is tricky: the distribution of noise on the median under the perturbation plus sub-sampling is a {\it convolution} of pdf of median of sub-sampled gradients and the pdf of the noisy median conditioned on the sub-sampled gradients. When there are a finite number of data to sub-sample, pdf of median of sub-sampled gradients is a summation of shifted delta distribution. Thus the pdf of median is actually a shifted sum of the pdf of noisy median conditioned on sub-sampled gradients. One can use Taylor expansion on pdf to show that if pdf of noisy median conditioned on sub-sampled gradients is symmetric, the pdf of median will become increasingly symmetric as the variance grows. Combining with our current results saying that the asymmetric part of pdf of noisy median conditioned on sub-sampled gradients dimishes as variance grows. It could be proved that pdf of sample median is symmetric and it will approach mean as variance increases. With symmetricity on pdf, the convergence of \signSGD{} could be expected. 
\xcedit{Although more efforts are required to make the analyses rigorous, one could expect both \signSGD{} and \medianSGD{} work under  sub-sampling plus noise perturbation.}
\mhedit{Additionally, incorporating geometry of neural nets into the analysis could also provide more insights on how to set the noise level and sub-sampling adaptively.} 

\noindent{\bf Is perturbation mechanism Byzantine robust?} 
We give an intuitive analysis here. First of all, \medianSGD{} converges to stationary points of the population risk as the number of data grows, when data are drawn iid. When the data are heterogenous, due to the gap between median and mean of gradients, it also suffers from a convergence gap on population risk. The perturbation mechanism and sub-sampling could diminish this gap when there is no Byzantine workers. On the other hand, the strongest attack that Byzantine nodes can do for median-based algorithms, is to send  gradients with extreme values simultaneously, shifting the median to  $k$th ordered statistic ($k \in [2n+1]$ depending on number of Byzantine nodes) of the stochastic gradients from good workers. As the noise increases, the difference between median and $k$th order statistic will also increase which means the algorithms may be more vulnerable to Byzantine workers. Thus, there may be a trade-off between Byzantine robustness and the performance of the algorithm without Byzantine workers.

\noindent{\bf What is the best distribution for perturbation?} 
We found in experiments that Gaussian distribution, Laplace distribution and uniform distribution all works well, but analyzing the latter two is more difficult due to the non-differentiable points.

\noindent{\bf Other applications and connections to dither?} As mentioned, our perturbation mechanism is a general approach to decrease the difference between expected median and mean, it will be interesting to see more applications of the approach. Further, we found our perturbation mechanism share some similarity with "dither" \citep{wannamaker2000theory} which aims at reducing artifacts in quantized signals by adding noise before quantization. It will also be interesting to investigate deeper connections between dither and our perturbation mechanism.

\bibliography{ref}
\bibliographystyle{plainnat}

\newpage
\appendix

\section{Proof of Proposition \ref{prop: sign_med}}\label{app: proof_prop_1}
Suppose we have a set of numbers $a_k, k \in [M]$, $a_k \neq 0, \forall k$ and $M$ is odd. \mhedit{We show the following identity} 
\begin{align}
      \mathrm{sign} \left( \sum_{k=1}^M \,\mathrm{sign}(a_k)\right)   =  \mathrm{sign}( \mathrm{median}(\{a_k\}_{k=1}^M))
\end{align}
\mhdelete{where the equality is because of following facts.}
To begin with, define $b_k, k \in [M]$ to be a sequence of $a_k$ sorted in ascending order. Then we have
\begin{align}
    \mathrm{median}(\{a_k\}_{k=1}^M) = \mathrm{median}(\{b_k\}_{k=1}^M) = b_{(M+1)/2}
\end{align}
and the following
\begin{align}
    \mathrm{sign} \left( \sum_{k=1}^M \,\mathrm{sign}(a_k)\right) = & \mathrm{sign} \left( \sum_{k=1}^M \,\mathrm{sign}(b_k)\right) \nonumber \\
    = & \mathrm{sign} \left(\,\mathrm{sign}(b_{(M+1)/2})+ \sum_{k=1}^{(M+1)/2-1} \,\mathrm{sign}(b_k) + \sum_{k=(M+1)/2+1}^{M} \,\mathrm{sign}(b_k)  \right).
\end{align}
Recall that $b_{k}$ is non-decreasing as it is a sorted sequence of $a_k$ with ascending order. If $b_{(M+1)/2} > 0$, we have $b_{k} > 0, \forall k > (M+1)/2$ and thus
\begin{align}
\sum_{k=(M+1)/2+1}^{M} \,\mathrm{sign}(b_k) = \sum_{k=(M+1)/2+1}^{M} 1 = (M-1)/2.
\end{align} 
Since $\sum_{k=(M+1)/2+1}^{M} \,\mathrm{sign}(b_k) \geq \sum_{k=(M+1)/2+1}^{M} -1 = -(M-1)/2$, we have
\begin{align}
 \,\mathrm{sign}(b_{(M+1)/2})+ \sum_{k=1}^{(M+1)/2-1} \,\mathrm{sign}(b_k) + \sum_{k=(M+1)/2+1}^{M} \,\mathrm{sign}(b_k)  \geq \left(\,\mathrm{sign}(b_k)  \right) = 1
\end{align}
which means when $   \mathrm{median}(a_k)> 0$,
\begin{align}
     \mathrm{sign} \left( \sum_{k=1}^M \,\mathrm{sign}(a_k)\right) = 1 
\end{align}

Following the same procedures as above, one can also get when $  \mathrm{median}(a_k)< 0$,
\begin{align}
     \mathrm{sign} \left( \sum_{k=1}^M \,\mathrm{sign}(a_k)\right) = -1 
\end{align}

Thus, 
\begin{align}
     \mathrm{sign} \left( \sum_{k=1}^M \,\mathrm{sign}(a_k)\right) =  \mathrm{sign}\left( \mathrm{median}(a_k) \right)
\end{align}
when $\mathrm{median}(a_k) \neq 0$.

Applying the result above to each coordinate of the gradient vectors finishes the proof. 
\hfill $\square$

\section{Proof of Theorem \ref{thm: signsgd}}\label{app: proof_thm1}
Let us define: 
\begin{align}
    \mathrm{median}(\{g_t\}) \triangleq \mathrm{median}(\{g_{t,i}\}_{i=1}^M).
\end{align}
and
\begin{align}
    \mathrm{median}(\{\nabla f_t\}) \triangleq \mathrm{median}(\{\nabla f_i(x_t)\}_{i=1}^M).
\end{align}

 By A3, we have the following standard descent lemma in nonconvex optimization.
 \begin{align} \label{eq: descent}
     f(x_{t+1}) \leq f(x_t) + \langle \nabla f(x_t), x_{t+1} - x_t \rangle + \frac{L}{2} \|x_{t+1} - x_t\|^2 
 \end{align}

Substituting the update rule into \eqref{eq: descent}, we have the following series of inequalities 
\begin{align} 
    f(x_{t+1}) & \leq  f(x_t) - \delta \langle \nabla f(x_t), \mathrm{sign}(\mathrm{median}(\{g_t\})) \rangle + \frac{L}{2} \delta^2 d \nonumber  \\
     &=  f(x_t) - \delta \langle \mathbb E[\mathrm{median}(\{g_t\})], \mathrm{sign}(\mathrm{median}(\{g_t\})) \rangle  \nonumber \\
     & \quad +\delta \langle \mathbb E[\mathrm{median}(\{g_t\})] - \nabla f(x_t) , \mathrm{sign}(\mathrm{median}(\{g_t\})) \rangle +
     \frac{L}{2} \delta^2 d \nonumber \\
     &\leq   f(x_t) - \delta \| \mathbb E[\mathrm{median}(\{g_t\})] \|_1 + \delta  \|\mathbb E[\mathrm{median}(\{g_t\})] - \nabla f(x_t)\|_1 \nonumber \\
     & + 2 \delta \sum_{j=1}^d |\mathbb E[\mathrm{median}(\{g_t\})_j]| I[\mathrm{sign}(\mathrm{median}(\{g_t\})_j) \neq \mathrm{sign}(\mathbb E[\mathrm{median}(\{g_t\})_j]) ] \nonumber \\
     & + \frac{L}{2} \delta^2 d
\end{align}
where $\mathrm{median}(\{g_t\})_j$ is $j$th coodrinate of $\mathrm{median}(\{g_t\})$, and $I[\cdot]$ denotes the indicator function.

Taking expectation over all the randomness, we get
\begin{align} \label{eq: expect_descent}
    &\delta \mathbb E[ \| \mathbb E[\mathrm{median}(\{g_t\})] \|_1   \nonumber \\ 
     & \leq   \mathbb E[f(x_t)] - \mathbb E[f(x_{t+1})] + \delta \mathbb E[\|\mathbb E[\mathrm{median}(\{g_t\})]] - \nabla f(x_t)\|_1] \nonumber \\
     & + 2 \delta E\left[\sum_{j=1}^d |\mathbb E[\mathrm{median}(\{g_t\})_j]| P[\mathrm{sign}(\mathrm{median}(\{g_t\})_j) \neq \mathrm{sign}(\mathbb E[\mathrm{median}(\{g_t\})_j]) ]\right] \nonumber \\
     & + \frac{L}{2} \delta^2 d
\end{align}

Before we proceed, we analyze the error probability of sign 
\begin{align}
    P[\mathrm{sign}(\mathrm{median}(\{g_t\})_j) \neq \mathrm{sign}(\mathbb E[\mathrm{median}(\{g_t\})_j]) ]
\end{align}
This follows a similar analysis as in \signSGD{} paper.

By reparameterization, we can have $$\mathrm{median}(\{g_t\})_j = \mathbb E[\mathrm{median}(\{g_t\})_j] + \zeta_{t,j}$$
with $\mathbb E[\zeta_{t,j}] = 0$.

By Markov inequality and Jensen's inequality, we have
\begin{align} \label{eq: error_prob}
 &P[\mathrm{sign}(\mathrm{median}(\{g_t\})_j) \neq \mathrm{sign}(\mathbb E[\mathrm{median}(\{g_t\})_j])] \nonumber \\
 &\leq P[|\zeta_{t,j}| \geq \mathbb E[\mathrm{median}(\{g_t\})_j] ] \nonumber \\
 &\leq \frac{\mathbb E[|\zeta_{t,j}|]}{\mathbb E[\mathrm{median}(\{g_t\})_j]} \nonumber \\
 &\leq  \frac{\sqrt{\mathbb E[\zeta_{t,j}^2]}}{\mathbb E[\mathrm{median}(\{g_t\})_j]} 
 = \frac{\sigma_{m}}{\mathbb E[\mathrm{median}(\{g_t\})_j]}
\end{align}
where we assumed $\mathbb E[\zeta_{t,j}^2] \leq \sigma_m^2$.

Substitute \eqref{eq: error_prob} into \eqref{eq: expect_descent}, we get
\begin{align} \label{eq: expect_descent2}
    &\delta \mathbb E[ \| \mathbb E[\mathrm{median}(\{g_t\})] \|_1  ] \nonumber \\ 
     & \leq   \mathbb E[f(x_t)] - \mathbb E[f(x_{t+1})] + \delta \mathbb E[\|\mathbb E[\mathrm{median}(\{g_t\})]] - \nabla f(x_t)\|_1]  + 2\delta d \sigma_m  + \frac{L}{2} \delta^2 d.
\end{align}
Now we use standard approach to analyze convergence rate. 
Summing over $t$ from 1 to $T$ and divide both sides by $T\delta$, we get
\begin{align} \label{eq: telescope_stoc}
    &\frac{1}{T} \sum_{t=1}^T \mathbb E[ \| \mathbb E[\mathrm{median}(\{g_t\})] \|_1 ] \nonumber\\ 
     &\leq   \frac{D_f}{T\delta} + \frac{1}{T }\sum_{t=1}^T \mathbb E[\|\mathbb E[\mathrm{median}(\{g_t\})]] - \nabla f(x_t)\|_1] 
      + 2 d \sigma_m  + \frac{L}{2} \delta d
\end{align}
where here we defined $D_f \triangleq \mathbb E[f(x_1)] \min_{x} f(x)$

Now set $\delta = \frac{\sqrt{D_f}}{\sqrt{LdT}}$, we get
\begin{align} \label{eq: rate}
    &\frac{1}{T} \sum_{t=1}^T \mathbb E[ \| \mathbb E[\mathrm{median}(\{g_t\})] \|_1]\nonumber\\   
     &\leq \frac{3}{2} \frac{\sqrt{dLD_f}}{\sqrt{T}} + \frac{1}{T }\sum_{t=1}^T \mathbb E[\|\mathbb E[\mathrm{median}(\{g_t\})]] - \nabla f(x_t)\|_1]  + 2 d \sigma_m  
\end{align}

Going one step further, \mhedit{and use the triangular inqaulity}, we can easily bound the $\ell_1$ norm of the gradient as the following
\begin{align} 
    &\frac{1}{T} \sum_{t=1}^T \mathbb E[ \| \nabla f(x_t)\|_1]   
     \leq  \frac{3}{2} \frac{\sqrt{dLD_f}}{\sqrt{T}} + 2 \frac{1}{T }\sum_{t=1}^T \mathbb E[\|\mathbb E[\mathrm{median}(\{g_t\})]] - \nabla f(x_t)\|_1]  + 2 d \sigma_m  
\end{align}

\section{Proof of Theorem \ref{thm: median_gd}}\label{app: proof_thm2}

By the gradient Lipschitz continuity and the update rule, we have
\begin{align*}
     &f(x_{t+1}) \nonumber \\ \leq & f(x_t) - \delta \langle \nabla f(x_t),\mathrm{median}(\{g_t\})\rangle  +\frac{L}{2} \delta^2 \| \mathrm{median}(\{g_t\})\|^2 \nonumber \\
     = & f(x_t) - \delta \|\nabla f(x_t)\|^2 + \delta \langle \nabla f(x_t),\nabla f(x_t) - \mathrm{median}(\{g_t\})\rangle  +\frac{L}{2} \delta^2 \| \mathrm{median}(\{g_t\})\|^2 \nonumber \\
     \leq &f(x_t) - \delta \|\nabla f(x_t)\|^2 + \delta \langle \nabla f(x_t),\nabla f(x_t) - \mathrm{median}(\{g_t\})\rangle  \\
     & +\frac{3L}{2} \delta^2 (\| \mathrm{median}(\{g_t\}) - \mathbb E[ \mathrm{median}(\{g_t\})|x_t] \|^2 + \| \mathbb E[\mathrm{median}(\{g_t\})|x_t] - \nabla f(x_t)\|^2 + \|\nabla f(x_t)\|^2) \nonumber
\end{align*}

Taking expectation, we have
\begin{align}\label{eq: descent_median}
    &\mathbb E[f(x_{t+1})] - \mathbb E[f(x_t)] \nonumber \\
     \leq & - \delta \mathbb E[ \|\nabla f(x_t)\|^2] + \delta \mathbb E[\langle \nabla f(x_t),\nabla f(x_t) - \mathrm{median}(\{g_t\})\rangle] \nonumber \\
     & +\frac{3L}{2} \delta^2 \mathbb E[\| \mathrm{median}(\{g_t\}) - \mathbb E[ \mathrm{median}(\{g_t\})|x_t] \|^2]  \nonumber \\
     & +\frac{3L}{2} \delta^2 (\mathbb E[ \| \mathbb E[\mathrm{median}(\{g_t\})] - \nabla f(x_t)\|^2] + \mathbb E[\|\nabla f(x_t)\|^2])
\end{align}

In addition, we have
\begin{align}\label{eq: cond_exp}
    &\mathbb E[\langle \nabla f(x_t),\nabla f(x_t) - \mathrm{median}(\{g_t\})\rangle] \nonumber \\
    = & E_{x_t}[\langle \nabla f(x_t),\nabla f(x_t) - \mathbb E[\mathrm{median}(\{g_t\})|x_t]\rangle] \nonumber \\
    = & E_{x_t}[\langle \nabla f(x_t) - \mathbb E[\mathrm{median}(\{g_t\})|x_t] ,\nabla f(x_t) - \mathbb E[\mathrm{median}(\{g_t\})|x_t]\rangle]  \nonumber \\
    & + \langle \mathbb E[\mathrm{median}(\{g_t\})|x_t] ,\nabla f(x_t) - \mathbb E[\mathrm{median}(\{g_t\})|x_t] \nonumber \\
    \leq & \mathbb E[\| \mathbb E[\mathrm{median}(\{g_t\})|x_t]\|\|\nabla f(x_t) - \mathbb E[\mathrm{median}(\{g_t\})|x_t]\|] + \mathbb E[\|\nabla f(x_t) - \mathbb E[\mathrm{median}(\{g_t\})|x_t]\|^2]
\end{align}

Substitute \eqref{eq: cond_exp} into \eqref{eq: descent_median}, assuming $|\mathbb E[\mathrm{median}(\{g_t\})_j|x_t] - \nabla f(x_t)_j| \leq C$ and $ \mathbb E[\| \mathrm{median}(\{g_t\})_j - \mathbb E[ \mathrm{median}(\{g_t\})_j|x_t]\|^2] \leq \sigma_m^2  $, we have
\begin{align}\label{eq: median_telescope}
    &\mathbb E[f(x_{t+1})] - \mathbb E[f(x_t)] \nonumber \\
     \leq & - (\delta - \frac{3L}{2} \delta^2) \mathbb E[ \|\nabla f(x_t)\|^2] + \delta \mathbb E[\| \mathbb E[\mathrm{median}(\{g_t\})|x_t]\|\|\nabla f(x_t) - \mathbb E[\mathrm{median}(\{g_t\})|x_t]\|] \nonumber \\
     &+ \delta \mathbb E[\|\nabla f(x_t) - \mathbb E[\mathrm{median}(\{g_t\})|x_t]\|^2] \nonumber \\
     & +\frac{3L}{2} \delta^2 (\sigma_m^2 d + C^2d ).
\end{align}

Setting $\delta = \min(\frac{1}{\sqrt{Td}},\frac{1}{3L})$, telescope sum and divide both sides by $T(\delta -\frac{3L}{2} \delta^2)$, we have
\begin{align}\label{eq: final_median_gd}
    &\frac{1}{T}\sum_{t=1}^T  \mathbb E[ \|\nabla f(x_t)\|^2] \nonumber \\
     \leq & \frac{2\sqrt{d}}{\sqrt{T}} (\mathbb E[f(x_{1})] - \mathbb E[f(x_{T+1})]) +  \frac{2}{T}\sum_{t=1}^T \mathbb E[\| \mathbb E[\mathrm{median}(\{g_t\})|x_t]\|\|\nabla f(x_t) - \mathbb E[\mathrm{median}(\{g_t\})|x_t]\|] \nonumber \\
     &+ \frac{2}{T}\sum_{t=1}^T \mathbb E[\|\nabla f(x_t) - \mathbb E[\mathrm{median}(\{g_t\})|x_t]\|^2] +{3L} \frac{\sqrt{d}}{\sqrt{T}} (\sigma_m^2 + C^2  )
\end{align}
This completes the proof. \hfill $\square$

\section{Proof of Theorem \ref{thm: diverge_signsgd}}\label{app: proof_thm3}
In this section, we show that our analysis is tight, in the sense that the constant gap 
\begin{align}
    \frac{1}{T } \sum_{t=1}^T \mathbb E[\|\mathbb E[\mathrm{median}(\{g_t\})]] - \nabla f(x_t)\|_1]
\end{align}
does exist in practice.

Consider the following problem
\begin{align}\label{eq: counter_example}
    \min_{x \in \mathbb{R}} f(x) \triangleq \  \frac{1}{3}\sum_{i=1}^3 \frac{1}{2}(x-a_i)^2
\end{align}
with  $a_1 < a_2 < a_3$. \mhedit{In particular, $f_i(x) = \frac{1}{2}(x-a_i)^2$, so each local node has only one data point. Since the entire problem is deterministic, and the local gradient is also deterministic (i.e., no subsampling is available), we will drop the expectation below.} 

It is readily seen that the median of gradient is always $x-a_2$. \mhedit{Therefore running \signSGD{} on the above problem is equivalent to running \signSGD{} to minimize $\frac{1}{3}\sum_{i=1}^{3}\frac{1}{2}(x-a_2)^2$}. From the Theorem 1 in \citet{bernstein2018signsgd}, the \signSGD{} will converge to $x=a_2$ as $T$ goes to $\infty$ and $\delta = O(\frac{1}{\sqrt{T}})$). 

On the other hand, at the point $x=a_2$, the median of gradients \mhedit{$ \mathrm{median}(\{g_t\})$} is 0 but the gradient of $f(x)$ is given by
\begin{align}\label{eq: gap_converge}
   \nabla f(a_2) = \frac{1}{3} \sum_{i=1}^3(x - a_i) = \frac{1}{3} ( (a_2 - a_3) + (a_2 - a_1))
\end{align}

Recall that for this problem, we also have for any $x_t$,
\begin{align} \label{eq: median_grad_gap}
    &\|\mathbb E[\mathrm{median}(\{g_t\})] - \nabla f(x_t)\|_1 \nonumber \\
    & = \|\mhedit{\mathrm{median}(\{g_t\}) - \nabla f(x_t)\|_1} \nonumber \\
    & = \left|x_t-a_2 - \frac{1}{3} \sum_{i=1}^3(x_t-a_i)\right| = \left|\frac{1}{3}(2a_2 - a_1 - a_3)\right|.
\end{align}
Comparing \eqref{eq: gap_converge} and \eqref{eq: median_grad_gap}, we conclude that at a given point $x = a_2$ (for which the \signSGD{} will converge to),  we have 
\begin{align}\label{eq: final_gap}
    \|\nabla f(x)\|_1 = \frac{1}{T }\sum_{t=1}^T \mathbb E[\|\mathbb E[\mathrm{median}(\{g_t\})]] - \nabla f(x_t)\|_1] = \left|\frac{1}{3}(2a_2 - a_1 - a_3)\right|.
\end{align}
Substituting $a_1 = 0, a_2 = 1, a_3 = 5$ \mhedit{(which satisfies $a_1<a_2<a_3$ assumed the beginning)} into \eqref{eq: final_gap} finishes the proof for \signSGD{}.

  
  The proof for \medianSGD{} uses the same construction as the proof of Theorem \ref{thm: diverge_signsgd}, i.e. we consider the problem
\begin{align}
    \min_{x \in \mathbb{R}} f(x) \triangleq \  \frac{1}{3}\sum_{i=1}^3 \frac{1}{2}(x-a_i)^2
\end{align}
with  $a_1 < a_2 < a_3$. Then from the update rule of \medianSGD{}, it reduces to running gradient descent to minimize $\frac{1}{2}(x-a_2)^2$. From classical results on convergence of gradient descent, the algorithm will converge to $x=a_2$ with any stepsize $\delta < 2/L$. 

At the point $x=a_2$, the median of gradients is zero but $\nabla f(x)$ is 
\begin{align}
   \nabla f(a_2) = \frac{1}{3} \sum_{i=1}^3(x - a_i) = \frac{1}{3} ( (a_2 - a_3) + (a_2 - a_1)).
\end{align}
In addition, for any $x_t$, the gap between median and mean of gradients satisfy
\begin{align}
        &\|\mathbb E[\mathrm{median}(\{g_t\})] - \nabla f(x_t)\|^2 \nonumber \\
    = &\left|x_t-a_2 - \frac{1}{3} \sum_{i=1}^3(x_t-a_i)\right|^2 = \left|\frac{1}{3}(2a_2 - a_1 - a_3)\right|^2
\end{align}
Combining all above, we have for $x = a_2$, we get
\begin{align}
    \|\nabla f(x)\|^2 = \frac{1}{T }\sum_{t=1}^T \mathbb E[\|\mathbb E[\mathrm{median}(\{g_t\})]] - \nabla f(x_t)\|^2]
     = \left|\frac{1}{3}(2a_2 - a_1 - a_3)\right|^2.
\end{align}
Setting $a_1 = 0, a_2 = 1, a_3 = 5$ we get $|\frac{1}{3}(2a_2 - a_1 - a_3)|^2 = 1$ and the proof is finished. 

 \section{Proof of Theorem \ref{thm: gap_median_mean}}\label{app: proof_thm5}

\subsection{Proof for (a)}
  Assume we have a set of numbers $u_1,..,u_{2n+1}$. Given a symmetric and unimodal noise distribution with mean 0 and variance $1$, denote its pdf to be $h_0(z)$ and its cdf to be $H_0(z)$.  Draw $2n+1$ samples from the distribution $\xi_1,...,\xi_{2n+1}$. 
 
 Given a constant $b$, define random variable $\hat{u}_i = u_i + b\xi_i$.
 Define $\tilde{u} \triangleq \mathrm{median}(\{\hat{u}_i\}_{i=1}^{2n+1})$ and its pdf and cdf to be $h(z)$ and $H(z)$, respectively. Define $\bar{u} \triangleq \frac{1}{2n+1}\sum_{i=1}^{2n+1} u_i $.
 
  Denote the pdf and cdf of $\hat u_i$ to be $h_i(z,b)$ and $H_i(z,b)$. Since $\hat u_i = u_i + b \xi_i$ is a scaled and shifted version of $\xi_i$, given $
 \xi_i$ has pdf $h_0(z)$ and cdf $H_0(z)$, we know $h_i(z,b)= \frac{1}{b} h_0(\frac{ z-u_i}{b})$ and $H_i(z,b) = H_0(\frac{z-u_i}{b})$ from basic probability theory. In addition, from symmetricity of $h_0(z)$, we also have $1-H_0(z) = H_0(-z)$.
 
 Define pdf of $\tilde{u}$ to be $h(z,b)$, from order statistic, we know
\begin{align}\label{eq: pdf_med}
    h(z,b) = \sum_{i=1}^{2n+1} h_i(z,b) \sum_{S \in \mathcal{S}_i} \prod_{j \in S}  H_j(z,b)  \prod_{k \in [2n+1]\setminus \{i,S\}}(1- H_k(z,b))
\end{align}
where $\mathcal{S}_i$  is the set of all $n$-combinations of items from the set $[2n+1] \setminus i$.

To simplify notation, we write the pdf into a more compact form
\begin{align}
    h(z,b) = \sum_{i, \{J,K\} \in \mathcal S'_i}  h_i(z,b)  \prod_{j \in J}  H_j(z,b)  \prod_{k \in K}(1- H_k(z,b))
\end{align}
 where the set $\mathcal{S}_i'$ is the set of all possible $\{J,K\}$ with $J$ being a combination of $n$ items from $[2n+1] \setminus i$ and $K = [2n+1] \setminus \{J,i\}$ and $i \in [2n+1]$ is omitted.
 
 Then the expectation of median can be calculated as
 {\small 
 \begin{align*}
 &\mathbb E[\tilde u] \\
 =&\int_{-\infty}^{\infty} z \sum_{i, \{J,K\} \in \mathcal S'_i}  h_i(z,b)  \prod_{j \in J}  H_j(z,b)  \prod_{k \in K}(1- H_k(z,b)) dz \\
 =&\sum_{i, \{J,K\} \in \mathcal S'_i} \int_{-\infty}^{+\infty}  (bz+{u_i}) \frac{1}{b} h_0(z)\prod_{j \in J} H_0(z+\frac{u_i-u_j}{b})\prod_{k \in K} (1-H_0(z+\frac{u_i-u_k}{b})) b dz\\
  =&\sum_{i,\{J,K\} \in \mathcal{S}_i'} \int_{-\infty}^{+\infty}  (b z+u_i) h_0(z)\\
  &\prod_{j \in J} \left(H_0(z)+\frac{u_i-u_j}{b}h_0(z)+\frac{(u_i-u_j)^2}{2b^2}h_0'(z'_j)\right)\prod_{k \in K} \left(1-H_0(z)-\frac{u_i-u_k}{b}h_0(z)-\frac{(u_i-u_j)^2}{2b^2}h_0'(z'_k)\right)dz
\end{align*}}%
where the second inequality is due to a changed of variable from $z$ to $\frac{z-u_i}{b}$, the last inequality is due to Taylor expansion and $z_j'\in[z_j,z_j+\frac{u_i-u_j}{\sigma}],z_k'\in[z_k,z_k+\frac{u_i-u_k}{\sigma}]$. 

Now we consider terms with different order w.r.t $b$ after expanding the Taylor expansion.

First, we start with the terms that is multiplied by $b$, the summation of coefficients in front of these terms equals
\begin{align*}
\sum_{i,\{J,K\} \in \mathcal S_i'}\int_{-\infty}^{+\infty} zh_0(z)\prod_{j \in J} H_0(z)^n \prod_{k \in K} (1-H_0(z))^n dz = 0
\end{align*}
due to symmetricity of $f$ over 0.

Then we consider the terms that are not multiplied by $b$, the summation of their coefficients equals
\begin{align*}
    &\sum_{i,\{J,K\} \in \mathcal S_i'} (u_i-u_j)(\int_{-\infty}^{+\infty} z h_0(z)\prod_{j \in J} H_0(z)^{n-1} \prod_{k \in K} (1-H_0(z))^nh_0(z) dz) \\
    &- \sum_{i,\{J,K\} \in \mathcal S_i'}(u_i-u_j) (\int_{-\infty}^{+\infty} z h_0(z)\prod_{j \in J} H_0(z)^{n} \prod_{k \in K} (1-H_0(z))^{n-1}h_0(z) dz) \\
    &+\sum_{i,\{J,K\} \in \mathcal S_i'}u_i(\int_{-\infty}^{+\infty}  h_0(z)H_0(z)^n(1-H_0(z))^n dz)\\
    =&0+0+\sum_{i=1}^{2n+1}u_i \binom{2n}{n}\int_{-\infty}^{+\infty}  H_0(z)^n(1-H_0(z))^n dH_0(z)\\\
\end{align*}
due to the cancelling in the summation (i.e. $\sum_{i,\{J,K\} \in \mathcal S_i'} (u_i-u_j) = 0$).

Further, we have 
\begin{align*}
&\sum_{i=1}^{2n+1}u_i \binom{2n}{n}\int_{-\infty}^{+\infty}  H_0(z)^n(1-H_0(z))^n dH_0(z)\\
\stackrel{(a)}{=}&\sum_{i=1}^{2n+1}u_i \binom{2n}{n}\int_{0}^{1}  y^n(1-y)^n dy\\
    =&\sum_{i=1}^{2n+1}u_i \binom{2n}{n}\frac{1}{n+1}\int_{0}^{1}  (1-y)^n dy^{n+1}\\
    =&\sum_{i=1}^{2n+1}u_i \binom{2n}{n}\frac{1}{n+1}\left(-\int_{0}^{1}  y^{n+1} d(1-y)^n\right)\\
    \stackrel{(b)}{=}&\sum_{i=1}^{2n+1}u_i \binom{2n}{n}\frac{n}{n+1}\int_{0}^{1}  y^{n+1}(1-y)^{n-1} dy\\
    =&\cdots\\
    =&\sum_{i=1}^{2n+1}u_i \binom{2n}{n}\frac{n(n-1)\cdots 1}{(n+1)(n+2)\cdots 2n}\int_{0}^{1}  y^{2n} dy\\
    =&\sum_{i=1}^{2n}u_i\binom{2n}{n}\frac{n!n!}{(2n+1)!}=\frac{1}{2n+1}\sum_{i=1}^{2n+1}u_i
\end{align*}
where (a) is due to a change of variable from $H_0(z)$ to $y$ and the omitted steps are just repeating steps from $(a)$ to $(b)$.

In the last step, we consider the rest of the terms (terms multiplied by $1/b$ or higher order w.r.t. $1/b$). Since $h_0,h_0'$ are bounded, for any non-negative integer $p,q,k$, there exists a constant $c>0$ such that:
{\small 
\begin{align*}
    \left|\int_{-\infty}^{+\infty}zh_0(z)(H_0(z)^ph_0(z)^qh_0'(z')^k)dz\right|&\leq \int_{-\infty}^{+\infty}|z||h_0(z)||(H_0(z)^ph_0(z)^qh_0'(z')^k)|dz\\
    &\leq c\int_{-\infty}^{+\infty}|z|h_0(z)dz\\
    &= c(\int_{-1}^{+1}|z|h_0(z)dz+\int_{1}^{+\infty}|z|h_0(z)dz+\int_{-\infty}^{-1}|z|h_0(z)dz)\\
    &\leq c(\int_{-1}^{+1}h_0(z)dz+\int_{1}^{+\infty}z^2h_0(z)dz+\int_{-\infty}^{-1}z^2h_0(z)dz)\\
    &\leq c(\int_{-1}^{+1}h_0(z)dz+\int_{-\infty}^{+\infty}z^2h_0(z)dz)\\
    &\leq c(\int_{-1}^{+1}h_0(z)dz+1)\\
    &\leq c \text{\quad\quad[Here's another constant still denoted as $c$]}
\end{align*}}
And also
{\small 
\begin{align*}
    \left|\int_{-\infty}^{+\infty}h_0(z)(H_0(z)^ph_0(z)^qh_0'(z')^k)dz\right|\leq \int_{-\infty}^{+\infty}h_0(z)|H_0(z)^ph_0(z)^qh_0'(z')^k|dz\leq c' \int_{-\infty}^{+\infty}h_0(z)dz= c'
\end{align*}}%
for some constant $c'$.

Then the coefficient of rest of the terms are bounded by constant, and the order of them are at least $\mathcal{O}(\frac{1}{b})$. Therefore $|\mathbb E[\mathrm{median}(\{\hat{u}_i\}_{i=1}^{2n+1}) - \frac{1}{2n+1}\sum_{i=1}^{2n+1} u_i]|=\mathcal{O}(\frac{1}{b})$ which proves \eqref{eq: scale_gap}.


Now we compute the order of the variance of median($\hat{u_i}$) in terms of $b$
{\small 
\begin{align*}
    &\mathrm{Var}(\mathrm{median}(\hat{u}_i))\\
    =&\mathbb{E}[\mathrm{median}(\hat{u}_i)^2]-\mathbb{E}[\mathrm{median}(\hat{u}_i)]^2\\
    \leq&  \mathbb{E}[\mathrm{median}(\hat{u}_i)^2] \\
    =&\sum_{i,\{J,K\} \in \mathcal{S}_i'} \int_{-\infty}^{+\infty}  z^2 h_i(z)\prod_{j \in J} H_j(z)\prod_{k \in K} (1-H_k(z))dz\\
   =&\sum_{i,\{J,K\} \in \mathcal{S}_i} \int_{-\infty}^{+\infty}  (\sigma z+u_i)^2 h_0(z)\prod_{j \in J} H_0(z+\frac{u_i-u_j}{\sigma})\prod_{k \in K} (1-H_0(z+\frac{u_i-u_k}{\sigma}))dz\\
    =&\sum_{i,\{J,K\} \in \mathcal{S}_i} \int_{-\infty}^{+\infty}  (\sigma z+u_i)^2 h_0(z)  \times  \\
    &\prod_{j \in J} \left(H_0(z)+\frac{u_i-u_j}{\sigma}h_0(z)+\frac{(u_i-u_j)^2}{2\sigma^2}h_0'(z'_j)\right)\prod_{k \in K} \left(1-H_0(z)-\frac{u_i-u_k}{\sigma}h_0(z)-\frac{(u_i-u_j)^2}{\sigma^2}h_0'(z'_k)\right)dz
\end{align*}}
where $z_j'\in[z_j,z_j+\frac{u_i-u_j}{\sigma}],z_k'\in[z_k,z_k+\frac{u_i-u_k}{\sigma}]$. Similar to the analysis in computing order of gap between median and mean, we consider terms after expanding the multiple formula. Note that we similarly have:
\begin{align*}
    \left|\int_{-\infty}^{+\infty}z^2h_0(z)(H_0(z)^ph_0(z)^qh_0'(z')^k)dz\right|&\leq \int_{-\infty}^{+\infty}z^2h_0(z)|(H_0(z)^ph_0(z)^qh_0'(z')^k)|dz\\
    &\leq c\int_{-\infty}^{+\infty}z^2h_0(z)dz\\
    &= c
\end{align*}
Therefore, after expansion and integration, the coefficients of any order of $b$ are also bounded by constant. 
Since the order of the terms w.r.t $b$ are less than 2, we can conclude that the variance of $\mathrm{Median}(\hat{u_i})$ is of order $\mathcal{O}(b^2)$ which proves \eqref{eq: variance_scale}.

\subsection{Proof for (b)}

This key idea of the proof in part is similar to that for part (a). We use Taylor expansion to expand different terms in pdf of sample median and identify the coefficient in front terms with different order w.r.t. $b$. The difference is that instead of doing second order Taylor expansion on $H_0$, we also need to do it for $h_0$, thus requiring $h_0''$ to be uniformly bounded and absolutely integrable. In addition, not every higher order term is multiplied by $h_0(z)$, thus more efforts are required for bounding the integration of higher order terms.

First, by a change of variable (change $z$ to $ \frac{z - \bar u }{b}$), \eqref{eq: pdf_med} can be written as
  \begin{align}\label{eq: pdf_centered}
      h(\bar{u}+ bz,b) = \sum_{i=1}^{2n+1} \frac{1}{b} h_0(\frac{\bar{u} - u_i}{b} + z) \sum_{S \in \mathcal{S}_i} \prod_{j \in S}  H_0(\frac{ \bar{u} - u_j}{b} + z)  \prod_{k \in [2n+1]\setminus \{i,S\}} H_0(-\frac{ \bar{u} - u_k}{b} - z)
  \end{align}
  
Using the Taylor expansion on $f$, we further have
\begin{align} \label{eq: taylor1}
     h_0(\frac{\bar{u} - u_i}{b} + z) = h_0(z) + h_0'(z)  (\frac{\bar{u} - u_i}{b}) + \frac{h_0''({z_1})}{2} (\frac{\bar{u} - u_i}{b})^2
\end{align}
with $z_1 \in (z,\frac{\bar{u} - u_i}{b} +z)$ or $z_1 \in (\frac{\bar{u} - u_i}{b} +z ,z)$.
Similarly, we have
\begin{align} \label{eq: taylor2}
    H_0(\frac{ \bar{u} - u_j}{b} + z) = H_0(z) + h_0(z) (\frac{\bar{u} - u_j}{b}) + \frac{h_0'(z_2)}{2} (\frac{\bar{u} - u_j}{b}) ^2
\end{align}
and 
\begin{align} \label{eq: taylor3}
    H_0(-\frac{\bar{u} - u_k}{b} - z) = H_0(-z) - h_0(-z) (\frac{u_k - \bar{u} }{b}) -  \frac{h_0'(-z_3)}{2} (\frac{u_k - \bar{u} }{b})^2
\end{align}
where $z_2 \in (z,\frac{\bar{u} - u_j}{b} +z)$ or $z_2 \in (\frac{\bar{u} - u_j}{b} +z ,z)$, $z_3 \in (z,\frac{u_k- \bar{u} }{b} +z)$ or $z_3 \in (\frac{u_k -\bar{u} }{b} +z ,z)$.

Substituting \eqref{eq: taylor1}, \eqref{eq: taylor2}, and \eqref{eq: taylor3} into \eqref{eq: pdf_centered}, following similar argument as one can notice following facts. 

1. Summation of all zeroth order terms multiplied by $1/b$ is 
\begin{align}
 \frac{1}{b} \sum_{i=1}^{2n+1}  h_0(z) \sum_{S \in \mathcal{S}_i} \prod_{j \in S}  H_0(z)  \prod_{k \in [n]\setminus \{i,S\}} H_0(-z) 
\end{align}

2. All the terms multiplied by $1/b^2$ cancels with each other after summation due to the definition of $\bar{u}$. I.e.
\begin{align}
   \sum_{i=1}^{2n+1} \frac{1}{b} h_0'(\frac{\bar{u} - u_i}{b}) \sum_{S \in \mathcal{S}_i}   H_0(z)^n   H_0(- z)^n  = 0
\end{align}
\begin{align}
   \sum_{i=1}^{2n+1} \frac{1}{b} h_0(z)   \sum_{S \in \mathcal{S}_i} \sum_{j \in S} h_0(z)(\frac{\bar{u} - u_j}{b})  H_0(z)^{n-1}   H_0(- z)^n  = 0
\end{align}
\begin{align}
   \sum_{i=1}^{2n+1} \frac{1}{b} h_0(z)   \sum_{S \in \mathcal{S}_i}   H_0(z)^{n}   \sum_{k \in [n]\setminus \{i,S\}}  h_0(-z)(\frac{\bar{u} - u_k}{b})  H_0(- z)^{n-1}  = 0
\end{align}

3. Excluding the terms above, the rest of the terms are upper bounded by the order of $O(1/b^3)$.

Thus by another change of variable (change  $z$ to $\frac{z}{b}$), we have
\begin{align}
    h(\bar{u}+z,b)  = \frac{1}{b}g(\frac{z}{b}) + 
    \frac{1}{b}v(\frac{z}{b})
\end{align}
where 
\begin{align}
    g(z) =  \sum_{i=1}^{2n+1}  h_0(z) \sum_{S \in \mathcal{S}_i} \prod_{j \in S}  H_0(z)  \prod_{k \in [n]\setminus \{i,S\}} H_0(-z) 
\end{align}
which is the pdf of sample median of $2n+1$ iid draws from $h_0$ and it is symmetric around 0 .

Further, observe that when $h_0(z)$, $h_0'(z)$, and $h''(z)$ are all absolutely upper bounded and absolutely integrable, integration of absolute value of each high order term in $v(z)$ can be upper bounded in the order of $O(\frac{\max_i |\bar{u}- u_i|^2}{b^2})$. This is because each term in $v(z)$ is at least multiplied by $1/b^2$ and one of $h_0(z)$, $h_0(z)$, $h_0'(z)$, $h_0''(z_1)$, $h_0'(z_2)$ and $h_0'(-z_3)$ ($z_1$, $z_2$, $z_3$  appears through remainder terms of the Taylor' theorem). The terms multiplied by $h_0(z)$, $h_0(-z)$, or $h_0'(z)$ absolutely integrates into a constant. The terms multiplied only by the remainder terms in the integration are more tricky, one need to rewrite the remainder term into integral form and exchange the order of integration to prove that the term integrates the order of $O(1/b^2)$. We do this process for one term in the following and the others are omitted.
\begin{align}
    &\int_{-\infty}^{\infty}\frac{h_0''(z_1)}{2}(\frac{\bar{u} - u_i}{b})^2  H_0(z) H_0(-z) dx \nonumber \\
    \leq &  \int_{-\infty}^{\infty}\left | \frac{h_0''(z_1)}{2} (\frac{\bar{u} - u_i}{b})^2 \right| \|H_0\|_{\infty} \|H_0\|_{\infty} dx \nonumber \\
    = & \|H_0\|_{\infty} \|H_0\|_{\infty}  \int_{-\infty}^{\infty}\left |{\int_{x}^{x+\frac{\bar{u} - u_i}{b}}h_0''(t)} (t-x) dt\right| dx
\end{align}
where the equality holds because $\frac{h_0''(z_1)}{2} (\frac{\bar{u} - u_i}{b})^2$ is the remainder term of the Taylor expansion when approximating $z+ \frac{\bar{u} - u_i}{b}$ at $z$ and we changed the remainder term from the mean-value form to the integral form.

Without loss of generality, we assume $\bar{u} - u_i \geq 0 $ (the proof is similar when it is less than 0), then we get
\begin{align}
    &\int_{-\infty}^{\infty}\left |{\int_{x}^{x+\frac{\bar{u} - u_i}{b}}h_0''(t)} (t-x) dt\right| dx \nonumber \\
    \leq & \int_{-\infty}^{\infty} {\int_{x}^{x+\frac{\bar{u} - u_i}{b}}|h_0''(t)}| |(t-x)| dt dx \nonumber \\
     = & \int_{-\infty}^{\infty}{\int_{t - \frac{\bar{u} - u_i}{b} }^{t} \left |h_0''(t)\right|} |(t-x)| dx dt \nonumber \\
     = & \int_{-\infty}^{\infty}{\int_{t - \frac{\bar{u} - u_i}{b} }^{t} \left |h_0''(t)\right|} (t-x) dx dt \nonumber \\
     = & \int_{-\infty}^{\infty}{\int_{t - \frac{\bar{u} - u_i}{b} }^{t} \left |h_0''(t)\right|} (t-x) dx dt \nonumber \\
      = & (\frac{\bar{u} - u_i}{b})^2 \int_{-\infty}^{\infty} \left |h_0''(t)\right| dt
\end{align}
which is $(\frac{\bar{u} - u_i}{b})^2$ times a constant.

Thus, we have 
\begin{align}
    \int_{-\infty}^{\infty} \frac{1}{b}|v(\frac{z}{b})|dz = \int_{-\infty}^{\infty} \frac{1}{b}|v({z})|b dz =  O(\frac{\max_i |\bar{u}- u_i|^2}{b^2}) 
\end{align}
which completes this part. \hfill $\square$


  \section{Proof of Theorem \ref{thm: noisy_signsgd}}\label{app: proof_thm6}
  Use the fact that the noise on median is approximately unimodal and symmetric, one may prove that \signSGD{} can converge to a stationary point. With symmetric and unimodal noise, the bias in \signSGD{} can be alternatively viewed as a decrease of effective learning rate, thus slowing down the optimization instead of leading a constant bias. This proof formalizes this idea by characterizing the asymmetricity of the noise ($O(1/\sigma^2)$) and then  follow a sharp analysis for \signSGD{}. The key difference \mhedit{from Theorem \ref{thm: signsgd}} is taking care of the bias introduced by the difference between median and mean. 

  \xcedit{Let us recall: 
\begin{align}
    &\mathrm{median}(\{g_t\}) \triangleq \mathrm{median}(\{(g_{t,i}\}_{i=1}^M),\\
    &\mathrm{median}(\{\nabla f_t\}) \triangleq \mathrm{median}(\{\nabla f_i(x_t)\}_{i=1}^M).
\end{align}}
where \mhcomment{the notation below is inconsistent with Algorithm 3.}
\begin{align}\label{eq: noise}
    g_{t,i} = \nabla f_i(x_t) + b \xi_{t,i}
\end{align}
where $\xi_{t,i}$ is a $d$ dimensional random vector with each element drawn i.i.d. from $N(0,1)$

By \eqref{eq: descent}, we have the following series of inequalities 
{\small 
\begin{align} 
    &f(x_{t+1}) -  f(x_t)  \nonumber \\
    \leq& - \delta \langle \nabla f(x_t), \mathrm{sign}(\mathrm{median}(\{g_t\})) \rangle + \frac{L}{2} \delta^2 d \nonumber  \\
    = &  - \delta \sum_{j=1}^d |\nabla f(x_t)_j| (I[ \textrm{sign}{(\textrm{median}(\{g_t\})_j}) =  \textrm{sign}({\nabla f(x_t)_j}) ] -  I[ \textrm{sign}{(\textrm{median}(\{g_t\})_j}) \neq \textrm{sign}({\nabla f(x_t)_j}) ]) \nonumber \\ 
     &+ \frac{L}{2} \delta^2 d
\end{align}}
where $\mathrm{median}(\{g_t\})_j$ is $j$th coodrinate of $\mathrm{median}(\{g_t\})$, and $I[\cdot]$ denotes the indicator function.

Taking expectation over all the randomness, we get \mhcomment{one more step to show the inequality below?}
{\small
\begin{align} \label{eq: expect_descent_refined}
    &\mathbb E[f(x_{t+1})] - \mathbb E[f(x_t)] \nonumber \\ 
    \leq &  - \delta \mathbb E\left[\sum_{j=1}^d |\nabla f(x_t)_j|\left( P[\mathrm{sign}(\mathrm{median}(\{g_t\})_j) = \mathrm{sign}(\nabla f(x_t)_j)] - P[\mathrm{sign}(\mathrm{median}(\{g_t\})_j) \neq \mathrm{sign}(\nabla f(x_t)_j)] \right)\right] \nonumber\\
    &\quad + \frac{L}{2} \delta^2 d 
\end{align}
}

  Now we need a refined analysis on the error probability.  In specific, we need an sharp analysis on the following quantity 
   \begin{align}
       P[\mathrm{sign}(\mathrm{median}(\{g_t\})_j) = \mathrm{sign}(\nabla f(x_t)_j) ] - P[\mathrm{sign}(\mathrm{median}(\{g_t\})_j) \neq \mathrm{sign}(\nabla f(x_t)_j) ].
   \end{align}
   Using reparameterization, we can rewrite $\mathrm{median}(\{g_t\})$ as 
   \begin{align}
       \mathrm{median}(\{g_t\}) = \nabla f(x_t)+  \xi_{t} 
   \end{align}
   where $\xi_{t}$ is created by $\xi_{t,i}$'s added on the local gradients on different nodes. 
   
   Then, w.l.o.g., assume $\nabla f(x_t)_j \geq 0$ we have
   \begin{align}\label{eq: p_wrong}
& P[\mathrm{sign}(\mathrm{median}(\{g_t\})_j) \neq \nabla f(x_t)_j ] \nonumber \\
=& P [(\xi_t)_j \leq -  \nabla f(x_t)_j] \nonumber \\
=& \int_{-\infty}^{-\nabla f(x_t)_j} h_{t,j} (z)
   \end{align}
   where $h_{t,j} (z)$ is the pdf of the $j$th coordinate of $\xi_t$ .
  
   Similarly, we have
      \begin{align} \label{eq: p_correct}
& P[\mathrm{sign}(\mathrm{median}(\{g_t\})_j) = \nabla f(x_t)_j ] \nonumber \\
=& P [(\xi_t)_j > -  \nabla f(x_t)_j] \nonumber \\
=& \int_{-\nabla f(x_t)_j}^{\infty} h_{t,j} (z)
\end{align}

   From \eqref{eq: pdf_symmetric} and \eqref{eq:asymtc}, we can split $h_{t,j} (z)$ into a symmetric part and a non-symmetric part which can be written as \mhcomment{both should have zero mean, correct?}
   \begin{align}
       h_{t,j} (z) = h^s_{t,j} (z) + h^{
      u}_{t,j} (z)
   \end{align}
   where $h^s_{t,j} (z)$ is symmetric around 0 and $h^u_{t,j} (z)$ is not.
   
    Therefore, from \eqref{eq: p_correct} and \eqref{eq: p_wrong}, we know that 
      \begin{align}
       &P[\mathrm{sign}(\mathrm{median}(\{g_t\})_j) = \mathrm{sign}(\nabla f(x_t)_j) ] - P[\mathrm{sign}(\mathrm{median}(\{g_t\})_j) \neq \mathrm{sign}(\nabla f(x_t)_j) ] \nonumber \\
       = & \int_{-\nabla f(x_t)_j}^{\infty} h_{t,j} (z) - \int_{-\infty}^{-\nabla f(x_t)_j} h_{t,j} (z) \nonumber \\
       =& \int_{-\nabla f(x_t)_j}^{\infty} h^s_{t,j} (z) + h^{
      u}_{t,j} (z) - \int_{-\infty}^{-\nabla f(x_t)_j} h^s_{t,j} (z) + h^{
      u}_{t,j} (z) \nonumber \\
      = & \int_{-\nabla f(x_t)_j}^{\nabla f(x_t)_j} h^s_{t,j} (z) +  \int_{-\nabla f(x_t)_j}^{\infty}  h^{
      u}_{t,j} (z)  - \int_{-\infty}^{-\nabla f(x_t)_j}  h^{
      u}_{t,j} (z)
   \end{align}
   where the last equality is due to symmetricity of $h^s_{t,j} (z)$, \mhedit{and the assumption that $\nabla f(x_t)_j$ is positive}.
   
   To simplify the notations, define a new variable $z_{t,j}$ with pdf $h_{t,j}^s$, then we have
   \begin{align}
       \int_{-\nabla f(x_t)_j}^{\nabla f(x_t)_j} h^s_{t,j} (z) = P[|z_{t,j}| \leq |\nabla f(x_t)_j|]
   \end{align}
   A similar result can be derived for $\nabla f(x_t)_j\leq 0$.
   
   In addition, \xcedit{since the noise on each coordinate of local gradient satisfy Theorem \ref{thm: gap_median_mean}, apply \eqref{eq:asymtc} to each coordinate of the stochastic gradient vectors}, we know that \mhcomment{why? the connection is not clear}
   \begin{align}
       \int_{-\infty} ^{\infty} |h^u_{t,j} (z)| = O\left(\frac{\max_i |\nabla f_i(x_t)_j -\nabla f(x_t)_j|^2 }{b^2}\right)
   \end{align}
   and thus
   \begin{align}
       \int_{-\nabla f(x_t)_j}^{\infty}  h^{
      u}_{t,j} (z)  - \int_{-\infty}^{-\nabla f(x_t)_j}  h^{
      u}_{t,j} (z) = O\left(\frac{\max_i |\nabla f_i(x_t)_j -\nabla f(x_t)_j|^2 }{b^2}\right).
   \end{align}
  
 To continue, we need to introduce some new definitions.
    Define \mhcomment{maybe change to a different name since in (16) there is also S}
    \begin{align}\label{eq: snr}
        W_{t,j} = \frac{|\nabla f(x_t)_j|}{b\sigma_{mid}}
    \end{align}
    where $\sigma_{mid}$ is the variance of the noise with pdf \eqref{eq: pdf_symmetric}, that is
    \begin{align}
    g(z) =  \sum_{i=1}^{2n+1}  h_0(z) \sum_{S \in \mathcal{S}_i} \prod_{j \in S}  H_0(z)  \prod_{k \in [n]\setminus \{i,S\}} H_0(-z) 
\end{align}
    By adapting Lemma 1 from \citet{bernstein2019signsgd} (which is an application of Gauss's inequality), we have 
    \begin{equation}\label{eq: gauss}
        P[|z_{t,j}|<|\nabla f(x_t)_j|] \geq
        \begin{cases}  
        1/3 & W_{t,j} \geq \frac{2}{\sqrt{3}}\\
        \frac{W_{t,j}}{\sqrt{3}} & \mbox{otherwise}
        \end{cases}
    \end{equation}
    
Thus, continuing from \eqref{eq: expect_descent_refined}, we have \mhcomment{is the order in big O correct?}
{\small 
\begin{align} \label{eq: expect_descent_contd}
    &\mathbb E[f(x_{t+1})] - \mathbb E[f(x_t)] \nonumber \\ 
    \leq&   - \delta \mathbb E\left[\sum_{j=1}^d |\nabla f(x_t)_j|( P[\mathrm{sign}(\mathrm{median}(\{g_t\})_j) = \mathrm{sign}(\nabla f(x_t)_j)] - P[\mathrm{sign}(\mathrm{median}(\{g_t\})_j) \neq \mathrm{sign}(\nabla f(x_t)_j)] )\right] \nonumber \\
    &+ \frac{L}{2} \delta^2 d \nonumber \\
    \leq &  - \delta \mathbb E\left[\sum_{j=1}^d |\nabla f(x_t)_j|(P[|z_{t,j}|<|\nabla f(x_t)_j|])\right] \nonumber\\
    &\quad + \delta \mathbb E\left[\sum_{j=1}^d |\nabla f(x_t)_j|O\left(\frac{\max_i |\nabla f_i(x_t)_j -\nabla f(x_t)_j|^2 }{b^2}\right)\right] + \frac{L}{2} \delta^2 d 
\end{align}}%
Define $D_f \triangleq f(x_1) - \min_x{f(x)}$, telescope from 1 to $T$, set $\delta = 1/\sqrt{Td},\  b = T^{1/4}d^{1/4}$, divide both sides by $T\delta$, we have \mhcomment{is the order in big O correct?}\mhcomment{do we requires that the size of the gradient to be bounded? looks like the right hand side requires that?}
\begin{align} \label{eq: telescope}
    &\frac{1}{T}\sum_{t=1}^T  \mathbb E\left[\sum_{j=1}^d |\nabla f(x_t)_j|(P[|z_{t,j}|<|\nabla f(x_t)_j|])\right] \nonumber \\ 
    &\leq  \frac{\sqrt{d}}{\sqrt{T}}D_f  + \frac{1}{T}\sum_{t=1}^T  \mathbb E\left[\sum_{j=1}^d |\nabla f(x_t)_j|O\left(\frac{\max_i |\nabla f_i(x_t)_j -\nabla f(x_t)_j|^2 }{\sqrt{Td}}\right)\right] + \frac{L}{2} \frac{\sqrt{d}}{\sqrt{T}}  
\end{align}
where the RHS is decaying with a speed of $\frac{\sqrt{d}}{\sqrt{T}}$.

Further, substituting \eqref{eq: gauss} and \eqref{eq: snr} into \eqref{eq: telescope} 
 and multiplying both sides of \eqref{eq: telescope} by $3b\sigma_m$, we can get 
 \begin{align}
         &\frac{1}{T}\sum_{t=1}^T\left( \sum_{j \in \mathcal W_t}T^{1/4}d^{1/4}|\nabla f(x_t)_j| \sigma_m + \sum_{j \in [d] \setminus \mathcal W_t} \nabla f(x_t)_j^2\right) \nonumber \\ 
    &\leq  3 \sigma_m \frac{d^{3/4}}{T^{1/4}}D_f  + 3 \sigma_m \frac{1}{T}\sum_{t=1}^T  \mathbb E\left[\sum_{j=1}^d |\nabla f(x_t)_j|O\left(\frac{\max_i |\nabla f_i(x_t)_j -\nabla f(x_t)_j|^2 }{T^{1/4}d^{1/4}}\right)\right] +3 \sigma_m \frac{L}{2} \frac{d^{3/4}}{T^{1/4}} \nonumber \\
    &= O(\frac{d^{3/4}}{T^{1/4}})
 \end{align}
 which implies \eqref{eq: rate_signsgd_n} since $\sigma_m$ is a constant. 
 

 \hfill $\square$
\section{Proof of Theorem \ref{thm: noisy_median_gd}}\label{app: proof_thm7}

Following the same procedures as the Theorem \ref{thm: median_gd}, we can get
\begin{align}
    &\mathbb E[f(x_{t+1})] - \mathbb E[f(x_t)] \nonumber \\
     \leq & - (\delta-\frac{3L}{2} \delta^2) \mathbb E[ \|\nabla f(x_t)\|^2] + \delta \mathbb E[\| \mathbb E[\mathrm{median}(\{g_t\})|x_t]\|\|\nabla f(x_t) - \mathbb E[\mathrm{median}(\{g_t\})|x_t]\|] \nonumber \\
     &+ \delta \mathbb E[\|\nabla f(x_t) - \mathbb E[\mathrm{median}(\{g_t\})|x_t]\|^2] +\frac{3L}{2} \delta^2 (\sigma_m^2 d + C^2d )
\end{align}
which is the same as \eqref{eq: median_telescope}.

Sum over $t \in [T]$ and divide both sides by $T(\delta -\frac{3L}{2} \delta^2)$, assume $\delta \leq \frac{1}{3L}$, we get
\begin{align}
   &\frac{1}{T}\sum_{t=1}^T  \mathbb E[ \|\nabla f(x_t)\|^2] \nonumber \\
     \leq & \frac{2}{T\delta} (\mathbb E[f(x_{1})] - \mathbb E[f(x_{T+1})]) +  \frac{2}{T}\sum_{t=1}^T \mathbb E[\| \mathbb E[\mathrm{median}(\{g_t\})|x_t]\|\|\nabla f(x_t) - \mathbb E[\mathrm{median}(\{g_t\})|x_t]\|] \nonumber \\
     &+ \frac{2}{T}\sum_{t=1}^T \mathbb E[\|\nabla f(x_t) - \mathbb E[\mathrm{median}(\{g_t\})|x_t]\|^2] +{3Ld} \delta (\sigma_m^2 + C^2  )
\end{align}
By \eqref{eq: scale_gap} in Theorem \ref{thm: gap_median_mean}, we know \mhcomment{cite the exact result}\mhcomment{where is the $\sqrt{d}$ comes from?}
\begin{align}
    \mathbb E[\|\nabla f(x_t) - \mathbb E[\mathrm{median}(\{g_t\})|x_t]\| = O(\frac{\sqrt{d}}{b})
\end{align}
\xcedit{where $\sqrt{d}$ is due to $L_2$ norm.}
In addition, we have $\sigma_{m}^2 = O(b^2)$ by \eqref{eq: variance_scale} in Theorem \ref{thm: gap_median_mean}. \mhcomment{you mean Eq. (16)?}
Assume $ \mathbb E[\mathrm{median}(\{g_t\})_j|x_t] \leq Q$ and set $b = T^{1/4}d^{1/4}$ and $\delta = {T^{-3/4}d^{-3/4}}$, we get
\begin{align}
   &\frac{1}{T}\sum_{t=1}^T  \mathbb E[ \|\nabla f(x_t)\|^2] \nonumber \\
     \leq & \frac{2}{T \delta} (\mathbb E[f(x_{1})] - \mathbb E[f(x_{T+1})]) + O\left(\frac{d}{b}\right) + O\left(\frac{d}{b^2}\right) + O\left(\delta d b^2\right) \nonumber \\
     \leq & O\left(\frac{d^{3/4}}{T^{1/4}}+\frac{\sqrt{d}}{\sqrt{T}}\right) =  O\left(\frac{d^{3/4}}{T^{1/4}}\right).
\end{align}

\section{Details of the Implementation}
\label{Implementation_Detail}

Our experimentation is mainly implemented using Python 3.6.4 with packages MPI4Py 3.0.0, NumPy 1.14.2 and TensorFlow 1.10.0. We use the Message Passing Interface (MPI) to implement the distributed system, and use TensorFlow to implement the neural network. Up to 20 compute cores of two Intel Haswell E5-2680 CPUs with 64 GB of Memory were used for experiments to demonstrate the performance of the algorithm.

\subsection{Dataset and pre-processing} 
In the simulation, we use the MNIST dataset\footnote{Available at \href{http://yann.lecun.com/exdb/mnist/}{http://yann.lecun.com/exdb/mnist/}}, which contains a training set of 60,000 samples, and a test set of 10,000 samples, both are 28x28 grayscale images of the 10 handwritten digits. To facilitate the neural network training,  the original feature vector, which contains the integer pixel value from 0 to 255, has been scaled to a float vector in the range $(0, 1)$. The integer categorical label is also converted to the binary class matrix (one hot encoding) for use with the categorical cross-entropy loss.

\subsection{Neural Network and Initialization} 
A two-layer fully connected neural network with 128 and 10 neurons for each layer is used in the experiment. The initialization parameters are drawn from a truncated normal distribution centered on zero, with variance scaled with the number of input units in the weight tensor (fan-in).

\subsection{Parameter Tuning} 
We use constant stepsize and tuning the parameter from the set \{1, 0.1, 0.01, 0.001\}, in which 0.01 is used for Fig. \ref{fig:MNIST1} and 0.1 is used for Fig. \ref{fig:MNIST2}.

\end{document}